\documentclass{article} 
\usepackage{iclr2026_conference,times}

\usepackage{amsmath,amsfonts,bm}

\def\eqref#1{equation~\ref{#1}}

\def\1{\bm{1}}

\DeclareMathAlphabet{\mathsfit}{\encodingdefault}{\sfdefault}{m}{sl}
\SetMathAlphabet{\mathsfit}{bold}{\encodingdefault}{\sfdefault}{bx}{n}

\usepackage{hyperref}
\usepackage{url}
\usepackage{listings}
\usepackage{xcolor}
\usepackage{multirow}
\usepackage{graphicx}
\usepackage{makecell}
\usepackage{subcaption}
\usepackage{wrapfig}
\usepackage{booktabs}
\usepackage{amsmath,amssymb}
\usepackage{algorithm}
\usepackage{algpseudocode}
\usepackage{wrapfig}

\usepackage[most]{tcolorbox}
\usepackage{pgfplots}
\usetikzlibrary{pgfplots.groupplots}
\pgfplotsset{compat=1.3}
\usepackage{tikz}
\usetikzlibrary{patterns}

\newlength\savewidth

\definecolor{baselinecolor}{gray}{.9}

\definecolor{TrustOrange}{RGB}{217,119,58}
\definecolor{BaseGray}{RGB}{105,105,105}

\newcommand{\dmdgain}[1]{{\scriptsize\textcolor{TrustOrange}{#1}}}
\newcommand{\ourscell}[2]{\makecell{#1\\[-2pt]\dmdgain{#2}}}

\newcommand{\trustcomment}[1]{%
    \Comment{\textcolor{TrustOrange}{#1}}
}
\newcommand{\basecomment}[1]{%
    \Comment{\textcolor{BaseGray}{#1}}
}
\newcommand{\trustname}[1]{%
    \textcolor{TrustOrange}{\textbf{#1}}
}

\definecolor{lightgreen}{RGB}{220,245,220}
\definecolor{lightred}{RGB}{245,220,220}
\definecolor{cornellred}{rgb}{0.7, 0.11, 0.11}
\definecolor{cadmiumgreen}{rgb}{0.0, 0.42, 0.24}
\definecolor{aliceblue}{rgb}{0.91, 0.94, 0.97}
\definecolor{darkblue}{rgb}{0.83, 0.89, 0.97}
\definecolor{Red7}{rgb}{0.941, 0.243, 0.243}
\definecolor{Green7}{RGB}{55, 178, 77}
\definecolor{Blue9}{rgb}{0.098,0.3,0.9}

\hypersetup{
  linkcolor = cornellred,
  citecolor  = TrustOrange,
  colorlinks = true,
  urlcolor = Blue9
}

\title{Disagreement-Aware Distribution Matching Distillation for Few-Step Diffusion Models}

\author{%
  \begingroup
  \small
  \renewcommand{\arraystretch}{1.15}%
  \begin{tabular}{@{}l@{}}
   Zikai Zhou$^{1}$, Zhengpeng Xie$^{1}$, Shitong Shao$^{1}$, Lichen Bai$^{1}$, Tianhao Zhang$^{3}$ \\ Bojun Cheng$^{1}$, Zeke Xie$^{1,2}$\thanks{Corresponding author.} \\[0.2em]
    $^1$The Hong Kong University of Science and Technology (Guangzhou)\\
    $^2$ The Hong Kong University of Science and Technology\;
    $^3$Catnip AI\\[0.2em]
    {\tt\small zzhou440@connect.hkust-gz.edu.cn}
  \end{tabular}
  \endgroup
}

\iclrfinalcopy 
\begin{document}

\maketitle

\begin{abstract}
Distribution Matching Distillation~(DMD) enables strong few-step diffusion generation by matching the student distribution to a pretrained teacher. Existing improvements to DMD often augment its supervision or modify its optimization machinery. We ask a complementary question: \emph{can DMD diagnose its own updates using signals already produced during training?} DMD inherently computes the discrepancy between a pretrained teacher and a fake model tracking the student distribution, but conventionally uses it primarily to determine the distribution matching direction. We find that the magnitude of this teacher--fake disagreement provides additional diagnostic information. Across extensive experiments, large disagreement in the high-noise regime consistently coincides with substantially higher-energy DMD updates. This reveals a dual role of disagreement: it determines not only \emph{where} the student should move, but also provides an intrinsic signal of \emph{how aggressively} the update should be applied. Based on this observation, we introduce Disagreement-Aware Distribution Matching Distillation (DA-DMD), which activates only in the high-noise regime and continuously modulates the update strength according to teacher--fake disagreement. DA-DMD reuses quantities already available during DMD training and requires no additional model, evaluator, or inference-time modification. We show that it exactly recovers vanilla DMD outside the selected region, reduces update energy only through selected samples, and exhibits redescending influence under extreme disagreement. Across multiple diffusion backbones, DA-DMD consistently improves few-step generation at the same inference cost. On SD3.5-L, DA-DMD improves HPSv2 from $28.23$ to $30.93$, while achieving a $64.4\%$ average win rate over DMD2.
\end{abstract}

\section{Introduction}
\label{sec:introduction}

Diffusion models have become a dominant paradigm for high-fidelity image generation, but their iterative sampling procedure often requires many denoising evaluations at inference time~\citep{ho2020denoising,song2020denoising,karras2022elucidating,shao2026,shao2026efficientvideodiffusionmodels}. This has motivated extensive research on compressing pretrained diffusion models into few-step generators, including progressive distillation, consistency-based approaches, and distribution matching methods~\citep{salimans2022progressive,song2023consistency,luo2023latent,yin2024one,yin2024improved}. Among them, \emph{Distribution Matching Distillation}~(DMD) directly aligns the distribution induced by a few-step student with that of a pretrained diffusion teacher, using a frozen teacher and a fake model that tracks the evolving student distribution~\citep{yin2024one,yin2024improved}. This distribution-level formulation has enabled strong few-step generation across a range of diffusion backbones.

Recent work has improved DMD through better fake-model optimization, auxiliary objectives, reward-guided optimization, and adaptive teacher mechanisms~\citep{yin2024improved,luo2026tdm,bai2026optimizing,jiang2025distribution,wu2026diversity,wu2026qwenimageflashobjectivedesign,zhou2024score}. We take a complementary perspective: \emph{can DMD make better use of the information it already produces during training?} Every DMD update compares the predictions of a pretrained teacher and a fake model at the same noisy sample. This teacher--fake discrepancy determines the direction of distribution matching, but its magnitude varies substantially across samples and noise levels. We therefore investigate whether this native variation also carries information about the behavior of the resulting update.

\begin{figure*}[t]
    \centering
    \includegraphics[width=0.92\textwidth]{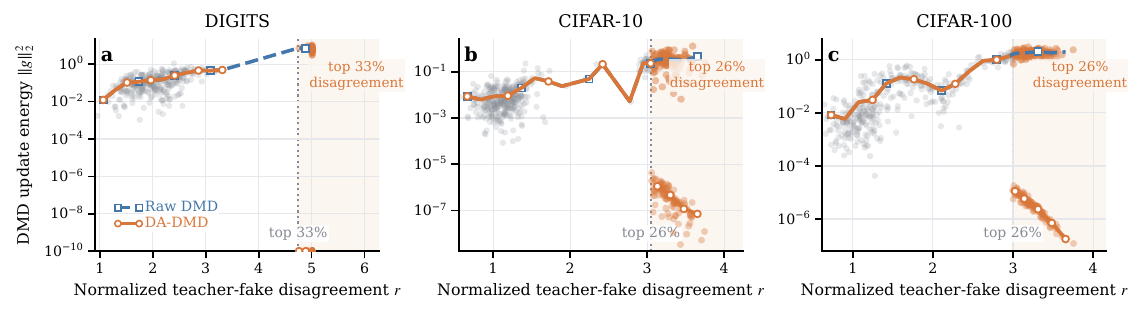}
    \vspace{-0.2in}
    \caption{
    \textbf{Teacher--fake disagreement diagnoses DMD update aggressiveness.}
    Within the predefined high-noise region $\mathcal H$, larger normalized disagreement is consistently associated with higher-energy raw DMD updates across DIGITS, CIFAR-10, and CIFAR-100.
    DA-DMD continuously attenuates these updates according to disagreement. Shaded regions highlight the high-disagreement tail for visualization only and do not correspond to an algorithmic threshold.
    }
    \label{fig:toy_training_gradient}
\end{figure*}

\begin{wrapfigure}[15]{r}{0.50\columnwidth}
    \vspace{-0.2in}
    \centering
    \includegraphics[width=\linewidth]
    {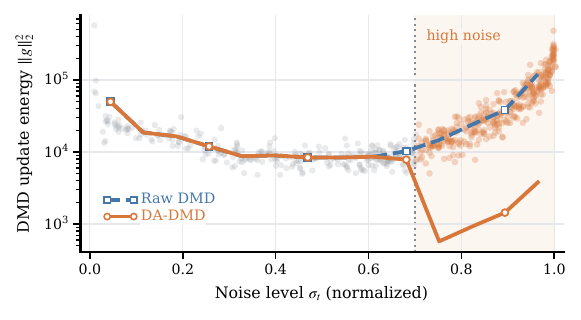}
    \vspace{-0.3in}
    \caption{
    \textbf{High-energy DMD updates concentrate at high noise.}
    During SD3.5-M distillation, raw DMD update energy rises sharply in the
    high-noise region $\mathcal H$, where DA-DMD activates disagreement-aware
    attenuation.
    }
    \label{fig:training_gradient}
    \vspace{-10pt}
\end{wrapfigure}

Our analysis reveals two complementary patterns. First, on real-world data from DIGITS~\citep{optical_recognition_of_handwritten_digits_80},
CIFAR-10~\citep{krizhevsky2009learning}, and CIFAR-100~\citep{krizhevsky2009learning}, larger normalized teacher--fake disagreement within the high-noise regime consistently coincides with higher-energy raw DMD updates
(Fig.~\ref{fig:toy_training_gradient}). Second, the same high-energy behavior persists at scale: during SD3.5-M~\citep{esser2024scaling} distillation, unusually high-energy raw DMD updates become increasingly concentrated in the
high-noise portion of the trajectory (Fig.~\ref{fig:training_gradient}). Importantly, these observations do not imply that high-noise or large-disagreement updates are inherently incorrect. Instead, together they suggest a simple optimization principle: \emph{use noise level to localize where aggressive DMD behavior tends to emerge, and use teacher--fake disagreement to continuously determine how strongly the update should be moderated.}

Motivated by these insights, we introduce \textbf{Disagreement-Aware Distribution Matching Distillation (DA-DMD)}. Rather than introducing a new source of guidance, DA-DMD reuses the teacher--fake discrepancy already computed by standard DMD. It activates only in a predefined high-noise region and continuously adjusts the DMD update strength according to normalized teacher--fake disagreement: updates outside this region remain exactly unchanged, while larger disagreement within the region induces stronger attenuation. DA-DMD therefore turns a native DMD signal that originally defines \emph{where} the student should move into an additional control signal for \emph{how conservatively} that move should be applied. Our ablations further show that applying the same correction globally or in lower-noise regions is less effective, supporting the selective high-noise design. 

DA-DMD is theoretically attractive because its robustness is selective rather than global. Outside the active high-noise region, it exactly recovers vanilla DMD, ensuring that ordinary distribution matching supervision is not weakened. Within the active region, disagreement-adaptive weighting reduces update energy only through the modified samples, while its redescending
influence prevents extreme teacher--fake disagreement from dominating optimization. Moreover, DA-DMD preserves the ideal fixed point of DMD, modifying the optimization dynamics rather than the target distribution. We conduct extensive experiments across multiple diffusion backbones,
including SD3, SD3.5~\citep{esser2024scaling}, SANA~\citep{xie2024sana}, and SANA-1.5~\citep{xie2025sana}, and evaluate on four complementary benchmarks: DrawBench~\citep{saharia2022photorealistic}, PartiPrompts~\citep{yu2022scaling}, GenEval2~\citep{kamath2025geneval}, and TIIF-Bench~\citep{wei2025tiif}. Across model families and evaluation suites, DA-DMD consistently improves the corresponding DMD baselines, with particularly clear gains on challenging compositional and fine-grained generation tasks. These improvements are achieved without changing the inference procedure
or sampling cost, while qualitative comparisons further demonstrate better visual fidelity.
Our contributions are threefold:
\begin{enumerate}
    \item  We uncover an overlooked role of teacher--fake disagreement, which contains previously underused diagnostic information: in the high-noise regime, larger disagreement is consistently associated with higher-energy DMD updates.

    \item We introduce Disagreement-Aware Distribution Matching Distillation (DA-DMD), which reuses this native signal to continuously modulate DMD updates in a predefined high-noise region while exactly preserving vanilla DMD elsewhere.

    \item We provide a theoretical analysis of disagreement-aware attenuation. Under standard smoothness and alignment assumptions, we show that adaptive attenuation yields a tighter one-step optimization bound when DMD update energy is sufficiently large relative to its descent signal, providing a principled explanation for selective intervention in the high-noise regime. We further establish exact preservation outside the active region and
    redescending influence under extreme disagreement, and extensively validate DA-DMD across multiple diffusion backbones and benchmarks.
\end{enumerate}

\section{Method}
\label{sec:method}

Our goal is to make better use of a signal that DMD already computes: the teacher--fake discrepancy. We first revisit how this discrepancy forms the standard DMD update. We then derive an adaptive reweighting from a simple teacher--fake mixture construction and turn it into a practical disagreement-aware update rule. Finally, we provide a theoretical analysis explaining when and why selective attenuation can be preferable to vanilla DMD.

\subsection{Distribution Matching Distillation}
\label{sec:dmd}

Distribution Matching Distillation (DMD)~\citep{yin2024one,yin2024improved} trains a few-step generator by matching the noisy marginal distribution induced by the student to that of a pretrained diffusion teacher. Let \(G_\theta\) denote the student generator and \(T\) the frozen teacher. Given condition \(c\) and initial noise \(z\), the student generates \(x_0^\theta=G_\theta(z,c)\), which is perturbed according to
\begin{equation}
\label{eq:forward_process}
x_t
=
\alpha_t x_0^\theta
+
\sigma_t\epsilon,
\qquad
\epsilon\sim\mathcal N(0,I).
\end{equation}
We denote the resulting student marginal by \(q_{\theta,t}(x_t\mid c)\) and the corresponding teacher marginal by \(p_{T,t}(x_t\mid c)\). DMD minimizes
\begin{equation}
\label{eq:dmd_objective}
\mathcal L_{\mathrm{DMD}}(\theta)
=
\mathbb E_{t,c}
\left[
\lambda_t
D_{\mathrm{KL}}
\left(
q_{\theta,t}(\cdot\mid c)
\,\Vert\,
p_{T,t}(\cdot\mid c)
\right)
\right],
\end{equation}
where \(\lambda_t\) is a timestep-dependent weighting coefficient.

The score of the evolving student marginal is not directly available, so DMD maintains a fake model \(F_\phi\) trained on student-generated samples to track the current student distribution. Let \(p_{F,t}\) denote the noisy marginal represented by the fake model, and define
\(s_T=\nabla_{x_t}\log p_{T,t}(x_t\mid c)\) and
\(s_F=\nabla_{x_t}\log p_{F,t}(x_t\mid c)\)
as the teacher and fake scores, respectively. Absorbing timestep-dependent coefficients into \(w_t\), the sample-wise DMD update can be written as
\begin{equation}
\label{eq:dmd_score_update}
g_{\mathrm{DMD}}
=
w_t
J_\theta^\top
\left(
s_F-s_T
\right),
\end{equation}
where \(J_\theta=\partial x_0^\theta/\partial\theta\).

To connect Eq.~\ref{eq:dmd_score_update} with the clean predictions directly produced by diffusion models, we use the Gaussian perturbation identity
\begin{equation}
\label{eq:score_x0_identity}
s_t(x_t,t,c)
=
\frac{
\alpha_t\mathbb E[x_0\mid x_t,c]-x_t
}{
\sigma_t^2
}.
\end{equation}
For an \(x_0\)-prediction model trained with squared-error denoising,
\(\hat{x}_0(x_t,t,c)\approx\mathbb E[x_0\mid x_t,c]\), and therefore
\begin{equation}
\label{eq:score_x0_prediction}
s_t(x_t,t,c)
\approx
\frac{
\alpha_t\hat{x}_0(x_t,t,c)-x_t
}{
\sigma_t^2
}.
\end{equation}
A complete derivation of Eq.~\ref{eq:score_x0_identity} from Fisher's identity is provided in Appendix~\ref{app:score_derivation}.

Since the teacher and fake model are evaluated at the same noisy sample \(x_t\), the shared term \(-x_t/\sigma_t^2\) cancels, giving
\begin{equation}
\label{eq:teacher_fake_score}
s_F-s_T
\approx
\frac{\alpha_t}{\sigma_t^2}
\left(
\hat{x}_{0,F}-\hat{x}_{0,T}
\right).
\end{equation}
Substituting Eq.~\ref{eq:teacher_fake_score} into Eq.~\ref{eq:dmd_score_update} yields
\begin{equation}
\label{eq:dmd_x0_update}
g_{\mathrm{DMD}}
\approx
\gamma_t
J_\theta^\top
\left(
\hat{x}_{0,F}-\hat{x}_{0,T}
\right),
\end{equation}
where \(\gamma_t=w_t\alpha_t/\sigma_t^2\) collects the timestep- and objective-dependent scalar factors.

Eq.~\ref{eq:dmd_x0_update} exposes the central signal driving DMD: the discrepancy between the teacher and fake predictions. Its direction determines the orientation of the distribution-matching correction, while its magnitude directly affects the resulting update strength. Importantly, this discrepancy varies substantially across samples and noise levels, as observed in Sec.~\ref{sec:introduction}. This raises a complementary question: \emph{can the teacher--fake discrepancy be used not only to construct the DMD correction, but also to determine how strongly that correction should be applied?}

\subsection{From Distribution Mixture to Disagreement-Aware DMD}
\label{sec:da_dmd}

\begin{figure*}[t]
\begin{center}
\includegraphics[width=.95\linewidth]{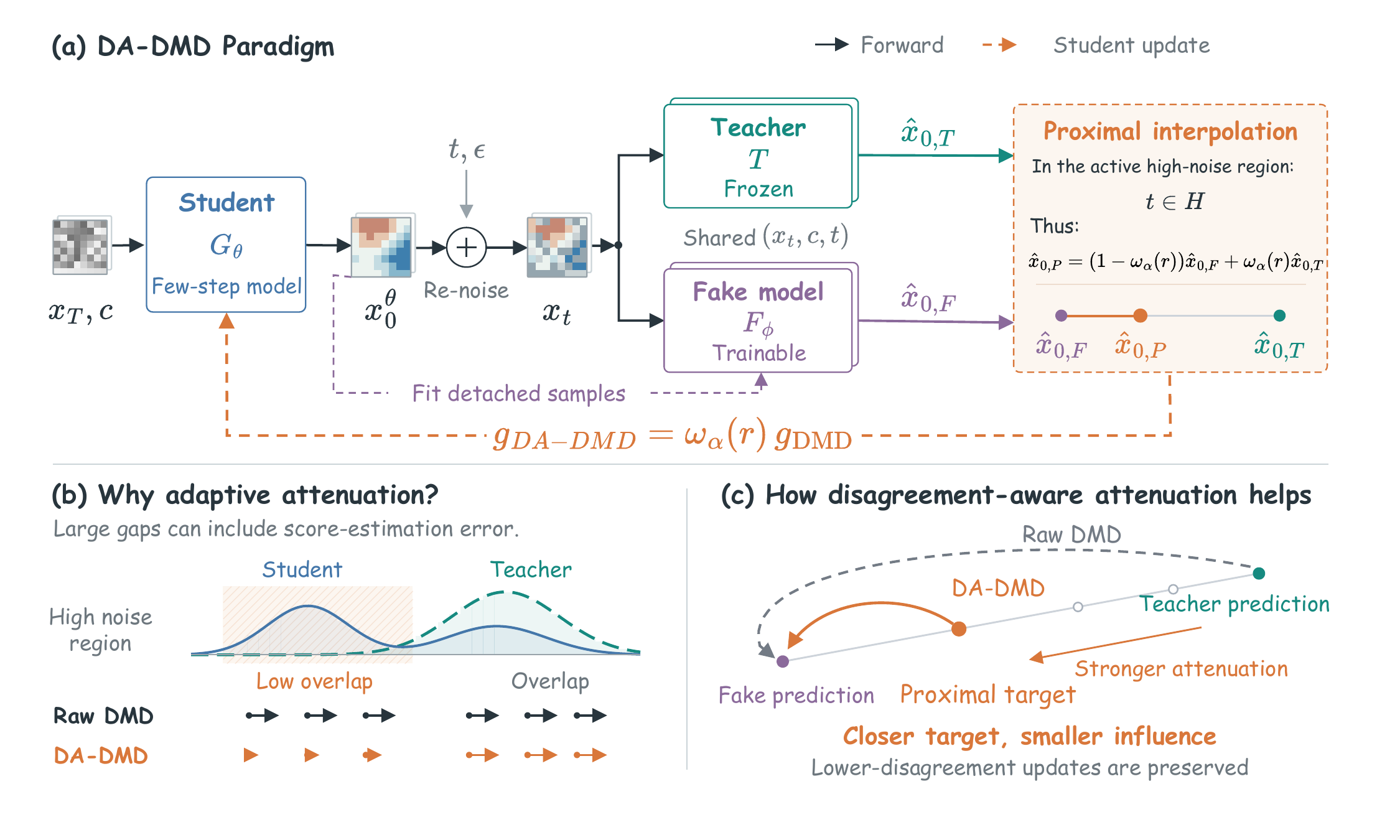}
\vspace{-0.3in}
\caption{
\textbf{Overview of DA-DMD.}
(a) In the active high-noise region, DA-DMD replaces the raw teacher-directed correction with a disagreement-aware proximal interpolation between the fake and teacher predictions, yielding an attenuated DMD update with multiplier $\omega_\alpha(r)$.
(b) Large teacher--fake disagreement can induce disproportionately large raw DMD corrections.
(c) DA-DMD shortens these corrections toward an adaptive proximal target, reducing the influence of high-disagreement updates while preserving lower-disagreement updates.
}
\label{fig:method}
\end{center}
\end{figure*}

\paragraph{From distribution mixture to adaptive DMD.}
To derive an adaptive DMD correction, we begin with a simple auxiliary mixture of the teacher and fake distributions. For clarity, we omit the conditioning variable \(c\). At noise level \(t\), define
\begin{equation}
\label{eq:mixture_distribution}
p_{\alpha,t}(x_t)
=
\alpha p_{T,t}(x_t)
+
(1-\alpha)p_{F,t}(x_t),
\qquad
\alpha\in(0,1).
\end{equation}
Although \(\alpha\) is fixed at the distribution level, the corresponding score mixture is sample dependent. Let
\(
\Delta_t(x_t)
=
\log \frac{p_{T,t}(x_t)}{p_{F,t}(x_t)}
\),
and define the generic mixture weighting function
\begin{equation}
\label{eq:mixture_weight_function}
\mathcal W_\alpha(u)
=
\frac{
\alpha e^u
}{
\alpha e^u+1-\alpha
}.
\end{equation}
Using
\(
p_{\alpha,t}(x_t)
=
p_{F,t}(x_t)
[\alpha e^{\Delta_t(x_t)}+1-\alpha]
\)
and
\(
\nabla_{x_t}\Delta_t=s_T-s_F
\),
the score of the auxiliary mixture becomes
\begin{equation}
\label{eq:adaptive_score_mixture}
s_\alpha
=
\mathcal W_\alpha(\Delta_t)s_T
+
\left[
1-\mathcal W_\alpha(\Delta_t)
\right]s_F.
\end{equation}
Therefore,
\(
s_F-s_\alpha
=
\mathcal W_\alpha(\Delta_t)(s_F-s_T)
\),
which induces the adaptive DMD correction
\begin{equation}
\label{eq:adaptive_dmd_exact}
g_\alpha
=
\mathcal W_\alpha(\Delta_t)
g_{\mathrm{DMD}}.
\end{equation}
This sample adaptivity is specific to the arithmetic mixture. By contrast, the geometric interpolation
\(p_{\alpha,t}^{\mathrm{geo}}\propto p_{T,t}^{\alpha}p_{F,t}^{1-\alpha}\)
yields
\(s_{\alpha}^{\mathrm{geo}}=\alpha s_T+(1-\alpha)s_F\),
and therefore only a uniform rescaling of the DMD correction by \(\alpha\). We therefore adopt the arithmetic mixture to obtain sample-dependent weighting; a detailed comparison is provided in Appendix~\ref{app:mixture_comparison}.

\paragraph{From density ratio to native disagreement.}
The exact adaptive weight in Eq.~\ref{eq:adaptive_score_mixture} depends on the sample-wise log-density ratio \(\Delta_t(x_t)\), which is unavailable in standard diffusion models. We therefore instantiate the same weighting principle using a tractable signal already produced during DMD training. From Eq.~\ref{eq:teacher_fake_score}, the teacher--fake clean-prediction discrepancy is directly related to the corresponding score discrepancy. We define its noise-normalized magnitude as
\begin{equation}
\label{eq:normalized_disagreement}
r^2
=
\operatorname{mean}
\left[
\left(
\frac{
\hat{x}_{0,F}-\hat{x}_{0,T}
}{
\sigma_t
}
\right)^2
\right],
\end{equation}
where the mean is taken over the non-batch latent dimensions. By Eq.~\ref{eq:teacher_fake_score},
\(
(\hat{x}_{0,F}-\hat{x}_{0,T})/\sigma_t
\approx
(\sigma_t/\alpha_t)(s_F-s_T)
\),
so \(r\) measures the magnitude of the same native teacher--fake discrepancy that drives the DMD correction, after accounting for its noise-dependent scale.

To convert this discrepancy into an adaptive weight, we use the monotone quadratic-energy surrogate
\(
\widetilde{\Delta}(r)=-r^2/2
\)
for the inaccessible sample-wise density evidence. Importantly, \(\widetilde{\Delta}(r)\) is not intended to estimate the true log-density ratio \(\Delta_t(x_t)\); it provides a conservative, disagreement-dependent surrogate within the same mixture-weighting form. The practical DA-DMD weight is therefore
\begin{equation}
\label{eq:da_weight}
\omega_\alpha(r)
:=
\mathcal W_\alpha
\left(
-\frac{r^2}{2}
\right)
=
\frac{
\alpha e^{-r^2/2}
}{
\alpha e^{-r^2/2}
+
1-\alpha
}.
\end{equation}
The weight decreases monotonically with disagreement, satisfying
\(
\omega_\alpha(0)=\alpha
\)
and
\(
\omega_\alpha(r)\rightarrow0
\)
as \(r\rightarrow\infty\). Hence, the same coefficient \(\alpha\) that defines the auxiliary mixture also determines the maximum retained DMD update strength in the practical weighting rule, while disagreement introduces additional sample-adaptive attenuation.

\paragraph{Selective high-noise intervention.}
The analysis in Sec.~\ref{sec:introduction} shows that unusually high-energy DMD updates concentrate in the high-noise regime. We therefore restrict disagreement-aware attenuation to a predefined high-noise region \(\mathcal H\). Let
\(
\chi=\mathbf{1}[t\in\mathcal H]
\).
The final DA-DMD update is
\begin{equation}
\label{eq:da_dmd_update}
g_{\mathrm{DA-DMD}}
=
\operatorname{sg}
\left[
1-\chi+\chi\omega_\alpha(r)
\right]
g_{\mathrm{DMD}},
\end{equation}
where \(\operatorname{sg}[\cdot]\) denotes stop-gradient. Equivalently,
\[
g_{\mathrm{DA-DMD}}
=
\begin{cases}
g_{\mathrm{DMD}}, & t\notin\mathcal H,\\[3pt]
\omega_\alpha(r)g_{\mathrm{DMD}}, & t\in\mathcal H.
\end{cases}
\]
Eq.~\ref{eq:da_dmd_update} separates the roles of noise level and disagreement: the high-noise gate determines \emph{where} DA-DMD intervenes, while normalized teacher--fake disagreement determines \emph{how strongly} the update is attenuated. Appendix Fig.~\ref{fig:motivation_weight} visualizes this weighting behavior, including the exact preservation outside \(\mathcal H\) and the progressively stronger attenuation within the active region.

The resulting effect on optimization is illustrated in  Appendix Fig.~\ref{fig:motivation_cifar10}, and Fig.~\ref{fig:motivation_digits} and~\ref{fig:motivation_cifar100}. In the controlled example, high-disagreement samples are associated with larger raw DMD updates, whereas DA-DMD reduces their influence while retaining the underlying DMD correction direction. This provides an intuitive view of how disagreement-aware attenuation can prevent a small subset of aggressive updates from disproportionately shaping the optimization trajectory.

All quantities required by DA-DMD are already available during standard DMD training. In all experiments, both our DMD2 baseline and DA-DMD use only the score-based distribution-matching branch, without the real-data adversarial objective. DA-DMD therefore modifies only this DMD update through disagreement-aware reweighting, while the student rollout, teacher query, fake-model optimization, and inference procedure remain unchanged.

\begin{algorithm}[h]
\caption{\trustname{DA-DMD}}
\label{alg:da-dmd}
\begin{algorithmic}[1]

\Require Teacher $T$, student $G_\theta$, fake model $F_\phi$;
weighting coefficient $\alpha\in(0,1)$;
noise schedule $\{\sigma_t\}$;
high-noise region $\mathcal H$

\For{each training iteration}

    \State Obtain $x_0^\theta$, $x_t$, $\hat{x}_{0,T}$,
    $\hat{x}_{0,F}$, and $g_{\mathrm{DMD}}$ as in DMD2
    \basecomment{standard DMD}

    \State
    $r^2 \gets
    \operatorname{mean}\!\left[
    \left(
    \frac{\hat{x}_{0,F}-\hat{x}_{0,T}}
    {\sigma_t}
    \right)^2
    \right]$
    \trustcomment{measure normalized disagreement}

    \State
    $\chi \gets
    \mathbf{1}\!\left[t\in\mathcal H\right]$
    \trustcomment{activate in high-noise region}

    \State
    $\omega_\alpha(r) \gets
    \dfrac{
    \alpha e^{-r^2/2}
    }{
    \alpha e^{-r^2/2}+1-\alpha
    }$
    \trustcomment{compute adaptive retention}

    \State
    $\textcolor{TrustOrange}{g_{\mathrm{DA-DMD}}}
    \gets
    \operatorname{sg}\!\left[
    1-\chi+\chi\omega_\alpha(r)
    \right]
    g_{\mathrm{DMD}}$
    \trustcomment{selectively reweight DMD}

    \State Update $G_\theta$ with
    $\textcolor{TrustOrange}{g_{\mathrm{DA-DMD}}}$;
    update $F_\phi$ as in DMD2

\EndFor
\end{algorithmic}
\end{algorithm}

\subsection{DA-DMD as Selective Adaptive Step Control}
\label{sec:theory}

The mixture construction in Sec.~\ref{sec:da_dmd} explains how the adaptive weighting rule arises. We now turn to a complementary question: when is attenuating a DMD update actually preferable to applying it at full strength? Importantly, DA-DMD does not interpret high disagreement as evidence of an incorrect update. Instead, our analysis considers a subtler regime in which the update remains aligned with a useful descent direction but its magnitude becomes overly aggressive.

Let \(g=g_{\mathrm{DMD}}\), and consider a generic attenuated step \(\theta^{+}=\theta-\eta a g\), where \(a\in[0,1]\). Assume that the DMD objective \(\mathcal L(\theta)\) is \(\beta\)-smooth and that the conditional DMD update has positive descent alignment. Conditioned on noise level \(t\) and disagreement \(r\), define \(A(t,r)=\mathbb E[\langle\nabla\mathcal L(\theta),g\rangle\mid t,r]>0\) as the useful descent alignment and \(B(t,r)=\mathbb E[\|g\|^2\mid t,r]>0\) as the update energy. The standard smoothness inequality gives
\begin{equation}
\label{eq:one_step_bound}
\mathbb E
\left[
\mathcal L(\theta^{+})
\mid t,r
\right]
\le
\mathcal L(\theta)
-
\eta a A(t,r)
+
\frac{\beta\eta^2a^2}{2}B(t,r).
\end{equation}
Minimizing the right-hand side over \(a\in[0,1]\) yields
\begin{equation}
\label{eq:optimal_attenuation}
a^\star(t,r)
=
\min
\left\{
1,
\frac{
A(t,r)
}{
\beta\eta B(t,r)
}
\right\}.
\end{equation}

Eq.~\ref{eq:optimal_attenuation} exposes the key optimization principle behind DA-DMD: \emph{the preferred update strength decreases when update energy becomes large relative to useful descent alignment}. Thus, a DMD update can remain positively aligned with the descent direction while still being too aggressive to apply at full strength. In other words, a useful update direction does not necessarily imply an appropriate update magnitude. More explicitly, for any fixed attenuation factor \(a\in(0,1)\), the corresponding conditional one-step bound is tighter than that of vanilla DMD whenever \(B(t,r)/A(t,r)>2/[\beta\eta(1+a)]\). The complete derivation is provided in Appendix~\ref{app:proof_attenuation}.

The analysis further connects the empirical disagreement signal to adaptive step control. Within the active high-noise region \(\mathcal H\), suppose that the conditional energy-to-alignment ratio \(B(t,r)/A(t,r)\) is nondecreasing with disagreement \(r\). Eq.~\ref{eq:optimal_attenuation} then implies
\[
r\uparrow
\quad\Longrightarrow\quad
a^\star(t,r)\downarrow.
\]
Such behavior is consistent with the practical DA-DMD weight \(\omega_\alpha(r)\) in Eq.~\ref{eq:da_weight}, which assigns stronger attenuation to larger disagreement. Under this assumption, teacher--fake disagreement therefore serves as an intrinsic control signal that follows the same monotonic step-control principle suggested by the smoothness analysis.

DA-DMD further localizes this robustification to the region where aggressive updates empirically concentrate. Outside the active region \(\mathcal H\), vanilla DMD is recovered exactly, while any induced reduction in update energy is confined to the high-noise regime. In addition, the disagreement-dependent influence is redescending:
\[
t\notin\mathcal H
\;\Longrightarrow\;
g_{\mathrm{DA-DMD}}=g_{\mathrm{DMD}},
\qquad
\lim_{r\rightarrow\infty}
r\,\omega_\alpha(r)=0.
\]
Consequently, DA-DMD moderates aggressive high-noise updates without globally weakening the original distribution-matching signal, while progressively reducing the influence of extreme teacher--fake disagreement. Complete proofs of these properties, together with complementary bias--variance and proximal interpretations, are provided in Appendix~\ref{app:proof_properties}.

\vspace{-0.2in}
\section{Experiments}
\label{sec:experiments}

In this section, we conduct extensive experiments to validate DA-DMD. Following the framework developed in Secs.~\ref{sec:da_dmd} and~\ref{sec:theory}, these experiments primarily serve to verify the effectiveness of disagreement-aware update control and the rationality of our selective high-noise design across multiple diffusion backbones and text-to-image benchmarks.

\vspace{-0.1in}
\subsection{Experimental Setup}
\label{sec:exp_setup}

We evaluate DA-DMD on 5 text-to-image diffusion backbones, including SANA~\citep{xie2024sana}, SANA-1.5~\citep{xie2025sana}, SD3-M, SD3.5-M, and SD3.5-L~\citep{esser2024scaling}. Our evaluation covers 4 complementary benchmarks: DrawBench~\citep{saharia2022photorealistic}, PartiPrompts~\citep{yu2022scaling}, GenEval2~\citep{kamath2025geneval}, and TIIF-Bench~\citep{wei2025tiif}, spanning general text-to-image generation, compositional fidelity, and fine-grained instruction following. We report PickScore~\citep{kirstain2023pick}, ImageReward~\citep{xu2023imagereward}, HPSv2~\citep{wu2023human}, AES~\citep{AES}, VisualQuality-R1~\citep{wu2026visualquality}, and average pairwise win rate as complementary measures of preference alignment and visual quality. DMD2~\citep{yin2024improved} serves as our primary baseline, together with representative few-step distillation and acceleration methods where applicable. Detailed benchmark descriptions, evaluation protocols, model configurations, and implementation settings are provided in the Appendix Sec.~\ref{apd:benchmark_and_dataset}.

\begin{table*}[t]
\centering
\caption{
\textbf{Quantitative comparison on DrawBench.}
DA-DMD provides broad improvements over DMD2 across multiple diffusion backbones and evaluation metrics.
}
\vspace{-0.1in}
\label{tab:drawbench_main}
\small
\setlength{\tabcolsep}{4.7pt}
\renewcommand{\arraystretch}{1.14}
\resizebox{0.95\textwidth}{!}{
\begin{tabular}{llcccccc}
\toprule
\textbf{Backbone} & \textbf{Method} & \textbf{PickScore}$\uparrow$ & \textbf{ImageReward}$\uparrow$ & \textbf{HPSv2}$\uparrow$ & \textbf{AES}$\uparrow$ & \textbf{VQ-R1}$\uparrow$ & \textbf{Avg. WR}$\uparrow$ \\
\midrule
\multirow{2}{*}[-0.9ex]{SANA} & DMD2~\citep{yin2024improved} & 22.53 & 95.05 & 29.67 & 5.820 & 4.580 & - \\
\cmidrule(lr){2-7}
 & \textcolor{TrustOrange}{\textbf{DA-DMD (Ours)}} & \ourscell{\textbf{22.56}}{+0.03} & \ourscell{\textbf{101.58}}{+6.53} & \ourscell{\textbf{29.87}}{+0.20} & \ourscell{\textbf{5.945}}{+0.125} & \ourscell{\textbf{4.643}}{+0.063} & \textbf{56.0\%} \\
\midrule
\multirow{2}{*}[-0.9ex]{SANA-1.5} & DMD2~\citep{yin2024improved} & 22.53 & 98.57 & 29.06 & 5.858 & 4.525 & - \\
\cmidrule(lr){2-7}
 & \textcolor{TrustOrange}{\textbf{DA-DMD (Ours)}} & \ourscell{\textbf{22.58}}{+0.05} & \ourscell{\textbf{102.09}}{+3.52} & \ourscell{\textbf{29.35}}{+0.29} & \ourscell{\textbf{5.916}}{+0.058} & \ourscell{\textbf{4.549}}{+0.024} & \textbf{57.0\%} \\
\midrule
\multirow{6}{*}{SD3-M} & DMD2~\citep{yin2024improved} & 22.33 & 98.56 & \underline{30.00} & 5.463 & 4.534 & - \\
 & Hyper-SD3~\citep{ren2024hyper} & 21.99 & 67.67 & 26.30 & 5.305 & 4.292 & 28.8\% \\
 & SiD-DiT~\citep{zhou2024score} & 22.46 & 91.31 & 28.01 & 5.287 & 4.599 & 39.8\% \\
 & SiD-Adv~\citep{zhou2024score} & \underline{22.53} & 90.67 & 29.62 & \textbf{5.560} & 4.624 & 50.2\% \\
 & TDM~\citep{luo2025learning} & 22.49 & \underline{101.13} & 29.81 & 5.474 & \textbf{4.693} & \underline{52.2\%} \\
\cmidrule(lr){2-7}
 & \textbf{\textcolor{TrustOrange}{DA-DMD (Ours)}} & \ourscell{\textbf{22.56}}{+0.23} & \ourscell{\textbf{110.72}}{+12.16} & \ourscell{\textbf{30.39}}{+0.39} & \ourscell{\underline{5.556}}{+0.093} & \ourscell{\underline{4.649}}{+0.115} & \textbf{61.2\%} \\
\midrule
\multirow{6}{*}{SD3.5-M} & DMD2~\citep{yin2024improved} & 22.49 & \underline{97.82} & 28.70 & 5.366 & 4.647 & - \\
 & SiD-DiT~\citep{zhou2024score} & \textbf{22.57} & 91.15 & 28.92 & 5.463 & \underline{4.682} & 48.0\% \\
 & SiD-Adv~\citep{zhou2024score} & 22.50 & 93.75 & 28.97 & \textbf{5.574} & 4.558 & 48.8\% \\
 & DP-DMD~\citep{wu2026diversity} & 22.32 & 91.37 & 28.08 & 5.299 & 4.553 & 39.4\% \\
 & SenseFlow~\citep{ge2025senseflow} & 22.42 & 89.37 & \underline{29.43} & \underline{5.564} & 4.648 & \underline{56.8\%} \\
\cmidrule(lr){2-7}
 & \textbf{\textcolor{TrustOrange}{DA-DMD (Ours)}} & \ourscell{\underline{22.51}}{+0.02} & \ourscell{\textbf{104.20}}{+6.38} & \ourscell{\textbf{29.57}}{+0.87} & \ourscell{5.474}{+0.108} & \ourscell{\textbf{4.700}}{+0.053} & \textbf{59.5\%} \\
\midrule
\multirow{6}{*}{SD3.5-L} & DMD2~\citep{yin2024improved} & 22.56 & 94.33 & 28.23 & 5.378 & 4.550 & - \\
 & SiD-DiT~\citep{zhou2024score} & 22.67 & 101.77 & 29.35 & 5.543 & 4.566 & 53.8\% \\
 & SiD-Adv~\citep{zhou2024score} & 22.69 & \underline{112.92} & \underline{30.52} & \textbf{5.743} & \underline{4.638} & \textbf{65.6\%} \\
 & SD3.5-L Turbo~\citep{ADD} & 22.65 & 91.07 & 28.47 & \underline{5.631} & 4.608 & 54.4\% \\
 & SenseFlow~\citep{ge2025senseflow} & \underline{22.70} & 112.52 & 30.18 & 5.421 & \textbf{4.652} & 62.2\% \\
\cmidrule(lr){2-7}
 & \textbf{\textcolor{TrustOrange}{DA-DMD (Ours)}} & \ourscell{\textbf{22.73}}{+0.17} & \ourscell{\textbf{113.88}}{+19.55} & \ourscell{\textbf{30.93}}{+2.70} & \ourscell{5.506}{+0.128} & \ourscell{4.587}{+0.037} & \underline{64.4\%} \\
\bottomrule
\end{tabular}
}
\end{table*}

\subsection{Main Results}
\label{sec:main_results}

\paragraph{Quantitative comparison.}
Tab.~\ref{tab:drawbench_main} and Appendix Tab.~\ref{tab:parti_main} report the main results on DrawBench and PartiPrompts. Across all 5 backbones, DA-DMD consistently improves the corresponding DMD2 baseline on both benchmarks, demonstrating that the benefit of disagreement-aware update control generalizes across different model families and scales. The gains are particularly pronounced on the SD3/SD3.5 family. On DrawBench, DA-DMD achieves average pairwise win rates of $61.2\%$, $59.5\%$, and $64.4\%$ on SD3-M, SD3.5-M, and SD3.5-L, respectively, while improving multiple preference- and quality-oriented metrics. Similar trends are observed on PartiPrompts, where DA-DMD reaches $61.2\%$ and $61.4\%$ average win rates on SD3.5-M and SD3.5-L. Compared with representative few-step distillation and acceleration baselines, DA-DMD remains competitive or superior across most evaluation dimensions while preserving the original inference procedure and sampling cost. Additional results on GenEval2 and TIIF-Bench in the Appendix Tab.~\ref{tab:geneval_all_models},~\ref{tab:geneval_atoms},~\ref{tab:tiif_basic},~\ref{tab:tiif_advanced} further show that these improvements extend to compositional generation and fine-grained instruction following.

\begin{figure*}[h]
\begin{center}
\includegraphics[width=.9\linewidth]{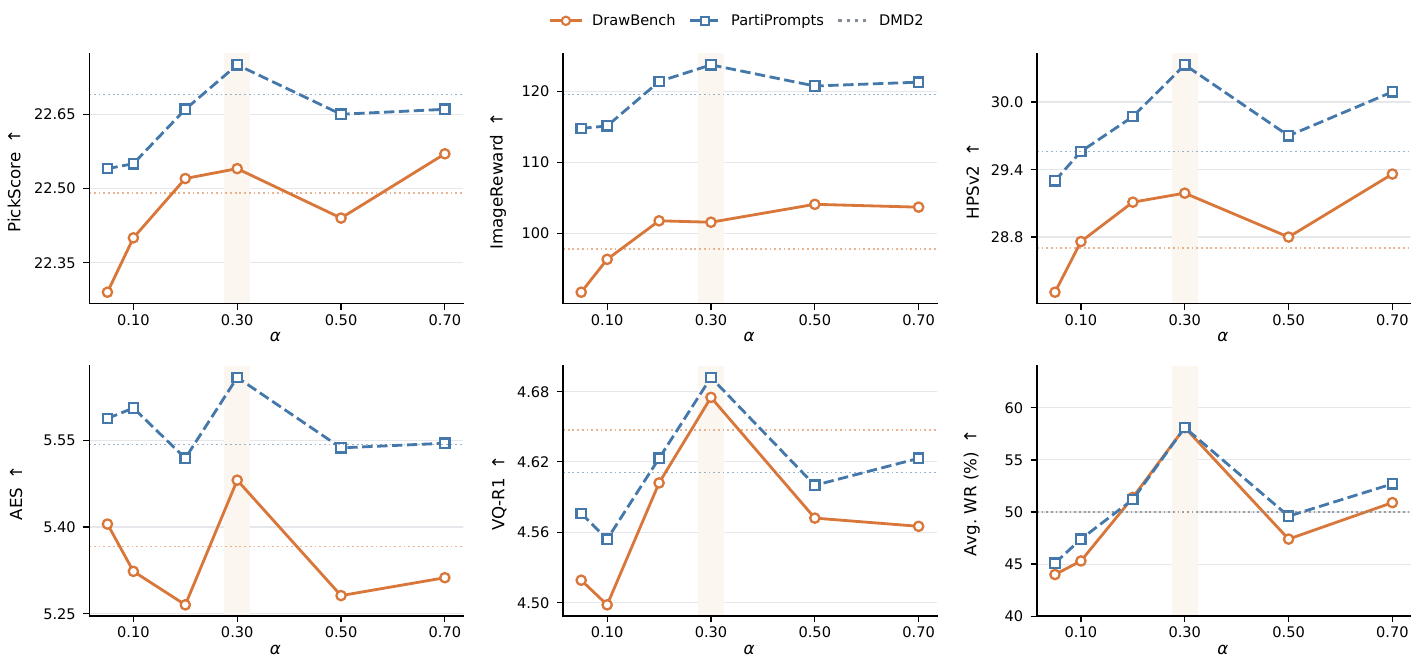}
\vspace{-0.15in}
\caption{
\textbf{Sensitivity to the weighting coefficient $\alpha$ on SD3.5-M.}
Moderate weighting, particularly $\alpha=0.3$, provides the best overall trade-off across metrics on DrawBench and PartiPrompts.
}
\label{fig:ablation_alpha}
\end{center}
\end{figure*}

\vspace{-0.25in}
\paragraph{Diagnosing DMD updates.}
Beyond generation quality, we examine whether teacher--fake disagreement provides a meaningful diagnostic of DMD optimization. Across DIGITS, CIFAR-10, and CIFAR-100, high-disagreement samples produce larger raw DMD updates and can disproportionately steer the optimization trajectory, whereas DA-DMD attenuates these corrections and recovers target-mode occupancy more faithfully (Figs. \ref{fig:motivation_cifar10}, ~\ref{fig:motivation_digits}, and~\ref{fig:motivation_cifar100}). Consistently, larger normalized disagreement within the high-noise regime coincides with higher-energy raw DMD updates across all three datasets (Fig.~\ref{fig:toy_training_gradient}). At model scale, SD3.5-M further shows that high-energy updates concentrate in the high-noise portion of the trajectory (Fig.~\ref{fig:training_gradient}). Together, these observations support the design of DA-DMD: noise level determines \emph{where} to intervene, while disagreement determines \emph{how strongly}.

\vspace{-0.1in}
\subsection{Ablation Study}
\label{sec:ablation}

\begin{figure*}[h]
\begin{center}
\includegraphics[width=.9\linewidth]{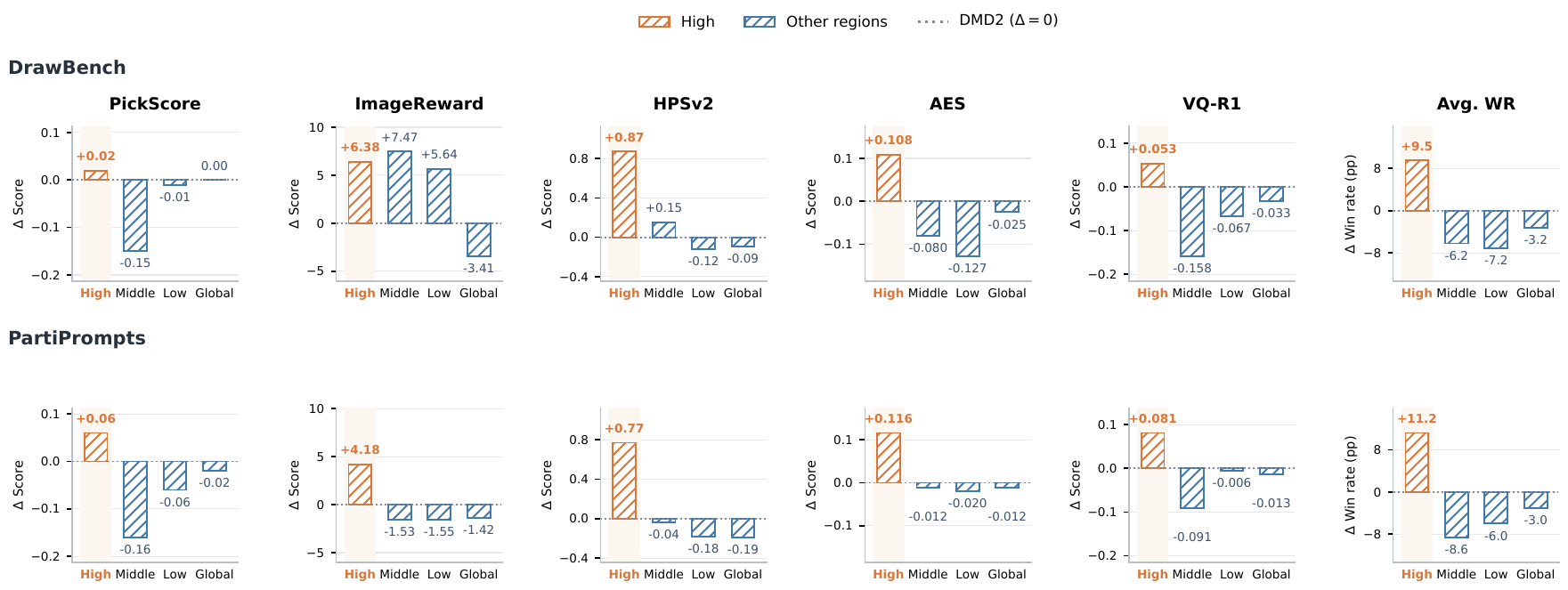}
\vspace{-0.1in}
\caption{
\textbf{Effect of the intervention region on SD3.5-M.}
High-noise intervention provides the strongest overall trade-off across DrawBench and PartiPrompts, while global, middle-noise, and low-noise attenuation are generally less effective.
}
\label{fig:ablation_region}
\end{center}
\end{figure*}

\paragraph{Where should DA-DMD intervene?}
We first investigate where disagreement-aware attenuation should be applied. As shown in Fig.~\ref{fig:ablation_region}, restricting DA-DMD to the high-noise region provides the strongest overall trade-off across DrawBench and PartiPrompts, with particularly clear gains in average pairwise win rate and most preference- and quality-oriented metrics. In contrast, global, middle-noise, and low-noise intervention is generally less effective and can underperform vanilla DMD on several metrics. These results indicate that disagreement-aware weighting is not uniformly beneficial across the diffusion trajectory, supporting our selective high-noise design.
\vspace{-0.1in}
\paragraph{Sensitivity to the weighting coefficient.}
We next study the effect of the weighting coefficient $\alpha$, which controls the maximum retained update strength in the active region through $\omega_\alpha(0)=\alpha$. Fig.~\ref{fig:ablation_alpha} shows that moderate weighting consistently provides the best overall trade-off across evaluation metrics, with $\alpha=0.3$ performing particularly well on both DrawBench and PartiPrompts. Smaller values tend to over-attenuate useful updates, whereas larger values provide weaker correction. Overall, DA-DMD exhibits a clear and interpretable sensitivity to $\alpha$, favoring moderate intervention over either extreme.

\vspace{-0.1in}
\subsection{Qualitative Results}
\label{sec:qualitative}

\begin{figure*}[t]
\begin{center}
\includegraphics[width=.9\linewidth]{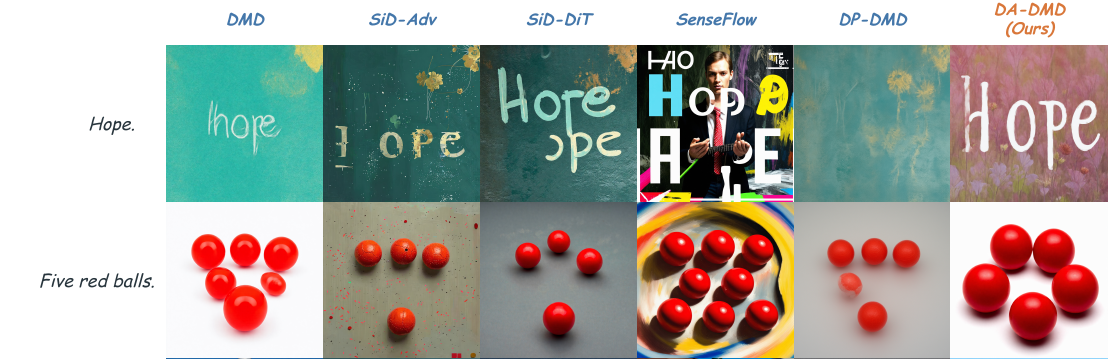}
\vspace{-0.1in}
\caption{
Qualitative comparison with representative distillation baselines on SD3.5-M.
}
\label{fig:sd35m_baseline_main}
\end{center}
\end{figure*}

Fig.~\ref{fig:sd35m_baseline_main} and Appendix Fig.~\ref{fig:sd35m_baseline} compare DA-DMD with representative few-step distillation and acceleration baselines on SD3.5-M. Across challenging prompts involving counting, text rendering, attribute binding, and object identity, DA-DMD produces generations that more faithfully satisfy the specified visual constraints while maintaining strong perceptual quality. These qualitative improvements are consistent with the quantitative gains reported in Sec.~\ref{sec:main_results}.

We further compare DA-DMD directly against DMD2 across all evaluated backbones. As shown in Appendix Fig.~\ref{fig:app:sd3m}, \ref{fig:app:sd35m}, \ref{fig:app:sd35l}, \ref{fig:app:sana}, and~\ref{fig:app:sana15}, DA-DMD consistently improves fine-grained prompt adherence across diverse compositional challenges, including object counting, spatial relations, relative attributes, color control, and structured multi-object generation. Together, these examples suggest that disagreement-aware update control improves not only automatic evaluation scores but also visually observable compositional fidelity.

\vspace{-0.2in}
\paragraph{Limitations.}
DA-DMD currently relies on a predefined high-noise intervention region and a hand-designed disagreement-to-weight mapping. While these choices are effective across the evaluated settings, the preferred intervention strength can vary across model families. A promising direction for future work is therefore to make both components more adaptive, allowing the model to determine when intervention is beneficial and how strongly the DMD correction should be modulated.
\vspace{-0.2in}
\paragraph{Conclusion.}
DA-DMD reuses the native teacher--fake discrepancy in DMD as an intrinsic signal for adaptive update control. Our key observation is that, beyond forming the distribution-matching correction itself, the magnitude of this discrepancy also provides useful information about the appropriate update strength, particularly in the high-noise regime where aggressive corrections tend to concentrate. Based on this observation, DA-DMD selectively attenuates high-noise updates according to disagreement while exactly preserving vanilla DMD outside the active region. Our analysis interprets this mechanism as adaptive step control, providing a principled explanation for why a directionally useful DMD correction can still benefit from a reduced step magnitude. Extensive experiments across multiple diffusion backbones and text-to-image benchmarks demonstrate broad improvements over DMD2 without introducing additional supervision or inference-time overhead. More broadly, our results suggest that native signals already present in distribution matching can be leveraged not only to construct the correction, but also to control how strongly it is applied.

\clearpage

\section*{AI Use Statement}
Generative AI tools were used to assist with research code implementation, manuscript organization, language polishing, and the presentation of mathematical derivations. All AI-assisted content was independently reviewed and verified by the authors, including manual inspection of generated code, mathematical arguments, experimental analyses, and manuscript revisions. The final research decisions, experimental design, technical claims, and scientific conclusions were made and validated by the authors, who take full responsibility for the content of this work.

\section*{Ethics Statement}
This work studies efficient distillation of text-to-image diffusion models and aims to reduce the computational cost of high-quality image generation. As with other generative modeling techniques, accelerated generation may inherit broader risks associated with image synthesis, including misuse for deceptive or inappropriate content. DA-DMD does not introduce new data collection, user profiling, or additional inference-time models, and is intended for responsible research and legitimate generative applications.

\section*{Reproducibility Statement}
We provide the formulation and training procedure of DA-DMD, including the disagreement measure, adaptive weighting rule, high-noise intervention strategy, and complete training pseudocode. We evaluate the method across multiple diffusion backbones and benchmarks, with experimental settings, evaluation protocols, ablations, and additional theoretical derivations provided in the Appendix. These details are intended to support reproducibility and facilitate comparison with existing DMD-based distillation methods.

\bibliography{iclr2026_conference}
\bibliographystyle{iclr2026_conference}


\appendix
\clearpage

\section{Implementation Details}
\label{apd:benchmark_and_dataset}

In this section, we provide an overview of the benchmarks, evaluation metrics, diffusion models and the hyperparameter settings in our main paper.

\subsection{Benchmark}
\label{app:benchmark}

\paragraph{DrawBench.} We evaluate general text-to-image generation capabilities using DrawBench~\citep{saharia2022photorealistic}, a challenging benchmark designed to probe a broad range of semantic and compositional skills. DrawBench contains 200 manually curated prompts organized into 11 categories, covering challenging aspects such as color and attribute binding, counting, spatial relations, text rendering, rare concepts, misspellings, and unusual object interactions. It further includes complex and long-form descriptions that require models to correctly compose multiple visual concepts. The diversity and compact size of DrawBench make it particularly suitable for fine-grained evaluation of both prompt alignment and perceptual generation quality.

\paragraph{PartiPrompts.} We further evaluate on PartiPrompts~(P2)~\citep{yu2022scaling}, a large-scale and diverse benchmark introduced for evaluating content-rich text-to-image generation. PartiPrompts consists of 1,632 English prompts spanning 12 semantic categories and 11 challenge aspects. The semantic categories cover a broad range of visual concepts, including people, animals, scenes, artifacts, abstract concepts, and world knowledge, while the challenge aspects range from basic generation and simple details to fine-grained attributes, quantity, spatial positioning, linguistic structures, writing and symbols, and complex compositions. This structured organization enables evaluation across both prompt content and generation difficulty, making PartiPrompts particularly useful for assessing generalization to diverse and compositionally demanding instructions.

\paragraph{TIIF-Bench.} To evaluate fine-grained instruction-following ability, we adopt TIIF-Bench~\citep{wei2025tiif}, which is specifically designed to assess how accurately modern text-to-image models interpret and execute complex user instructions. TIIF-Bench contains 5,000 diverse prompts organized across multiple dimensions and three levels of difficulty and complexity. For each instruction, the benchmark provides both a concise version and a rhetorically enriched long-form version with the same underlying semantics, allowing systematic evaluation of robustness to prompt length and linguistic complexity. Beyond conventional compositional requirements, TIIF-Bench also covers challenging capabilities such as text rendering, style control, and designer-level generation tasks. These characteristics make it well suited for measuring fine-grained instruction adherence under realistic and complex generation settings.

\paragraph{GenEval 2.} We additionally employ GenEval 2~\citep{kamath2025geneval}, an updated compositional text-to-image benchmark designed to address the increasing saturation and benchmark drift of earlier automatic evaluation suites. GenEval 2 contains 800 prompts with substantially improved coverage of primitive visual concepts and higher degrees of compositionality, including objects, attributes, spatial relations, and counting. Each prompt is decomposed into a set of atomic visual requirements, enabling evaluation not only at the prompt level but also at the level of individual compositional skills. GenEval 2 is evaluated using Soft-TIFA, a VQA-based evaluation framework that aggregates soft judgments over these visual primitives and is designed to better align automatic evaluation with human judgment. The benchmark therefore provides a fine-grained diagnostic of whether a model can simultaneously satisfy multiple visual constraints in increasingly complex prompts.

\subsection{Evaluation Metrics}
\label{app:evaluation_metrics}

\paragraph{PickScore.} PickScore~\citep{kirstain2023pick} is a CLIP-based human-preference metric trained on the Pick-a-Pic dataset, which contains pairwise preferences collected from real users of text-to-image systems. It jointly considers the input prompt and generated image to estimate human preference, providing an automatic measure of both semantic alignment and overall generation quality.

\paragraph{ImageReward.} ImageReward~\citep{xu2023imagereward} is a general-purpose text-to-image reward model trained on expert-annotated image comparisons. It is designed to capture human preferences over generated images while accounting for both text--image alignment and perceptual quality, and has been widely adopted for automatic evaluation of text-to-image generation.

\paragraph{HPS v2.} Human Preference Score v2 (HPS v2)~\citep{wu2023human} is a CLIP-based preference model trained on the Human Preference Dataset v2 (HPD v2), which contains large-scale pairwise human preference annotations. HPS v2 is designed to provide robust preference prediction across diverse image distributions and serves as a complementary measure of human-aligned generation quality.

\paragraph{AES.} We use the Aesthetic Score (AES)~\citep{AES} to evaluate the perceptual attractiveness of generated images. AES is predicted from CLIP image representations using an aesthetic predictor trained on human-rated images, providing a prompt-independent measure of visual aesthetics. Unlike preference metrics that jointly consider text and image, AES primarily reflects the overall aesthetic quality of the generated image.

\paragraph{VisualQuality-R1.} We further employ VisualQuality-R1~\citep{wu2026visualquality}, a reasoning-enhanced no-reference image quality assessment model trained with reinforcement learning to rank. VisualQuality-R1 evaluates perceptual image quality without requiring a reference image and provides a complementary assessment of visual fidelity beyond text--image preference metrics.

\subsection{Model Backbones}
\label{app:model_backbones}

We evaluate DA-DMD on five text-to-image diffusion backbones from two representative model families: SANA, SANA-1.5, Stable Diffusion 3 Medium (SD3-M), Stable Diffusion 3.5 Medium (SD3.5-M), and Stable Diffusion 3.5 Large (SD3.5-L). These models cover different architectures and model scales, allowing us to evaluate whether the proposed disagreement-aware optimization generalizes beyond a particular backbone.

\paragraph{SANA and SANA-1.5.} SANA~\citep{xie2024sana} is an efficient text-to-image framework based on a linear Diffusion Transformer architecture, designed to achieve high-resolution image generation with substantially reduced computational cost. SANA-1.5~\citep{xie2025sana} further scales the SANA family through improved training-time and inference-time scaling strategies. We include both generations of SANA to evaluate DA-DMD across closely related but differently scaled diffusion backbones.

\paragraph{Stable Diffusion 3 and 3.5.} Stable Diffusion 3~\citep{esser2024scaling} introduces the Multimodal Diffusion Transformer (MMDiT) architecture for text-to-image generation. We use SD3-M together with the subsequent SD3.5-M and SD3.5-L variants, covering both medium- and large-scale models. Evaluating DA-DMD across these three backbones allows us to study its behavior under different model capacities within the same general architecture family.

\subsection{Hyperparameter Settings}
\label{app:hyperparameters}

Unless otherwise specified, DA-DMD follows the same training recipe as the corresponding DMD2 baseline. Following the multi-step distribution-matching and fake-model optimization framework of DMD2, we use only the score-based distribution-matching objective in all experiments and disable the real-data adversarial objective for both DMD2 and DA-DMD. Therefore, the only difference between the two methods is the disagreement-aware reweighting applied to the DMD update.

All models are trained at $1024\times1024$ resolution with a 4-step student. The teacher is frozen, while both the student and fake model are initialized from the corresponding pretrained backbone. We use AdamW with a learning rate of $5\times10^{-6}$ for both the student and fake model, and perform five fake-model optimization steps per student update. For DA-DMD, the active high-noise region corresponds to the first $30\%$ of the noise trajectory, i.e., normalized noise level $\sigma_t^{\mathrm{norm}}\ge0.7$; outside this region, the original DMD update is left unchanged. The main model-specific settings are summarized in Tab.~\ref{tab:hyperparameters}.

\begin{table}[t]
\centering
\caption{
\textbf{Main training hyperparameters.}
DA-DMD uses the same optimization and sampling configuration as DMD2, with model-specific teacher guidance and disagreement weighting coefficient $\alpha$.
}
\label{tab:hyperparameters}
\small
\setlength{\tabcolsep}{5.0pt}
\renewcommand{\arraystretch}{1.12}
\begin{tabular}{lccc}
\toprule
\textbf{Backbone} &
\textbf{Teacher NFE} &
\textbf{Teacher CFG} &
\(\boldsymbol{\alpha}\) \\
\midrule
SANA        & 20 & 4.5 & 0.2 \\
SANA-1.5     & 20 & 4.5 & 0.2 \\
SD3-M       & 30 & 7.0 & 0.3 \\
SD3.5-M     & 30 & 3.5 & 0.3 \\
SD3.5-L     & 30 & 3.5 & 0.7 \\
\bottomrule
\end{tabular}
\end{table}

For all benchmarks, distilled models are evaluated using 4 sampling steps at $1024\times1024$ resolution with guidance scale $1.0$. We use the same evaluation seed and metric implementation for DMD2 and DA-DMD. The weighting coefficient $\alpha$ and intervention region are further analyzed in Sec.~\ref{sec:ablation}. The coefficient $\alpha$ controls the maximum retained DMD update strength within the active high-noise region through $\omega_\alpha(0)=\alpha$. We do not assume a universal value across model families. Instead, $\alpha$ is selected independently for each backbone from a small candidate set using the same tuning protocol. Our ablations show that performance is sensitive to the amount of attenuation and that moderate retention generally performs well, while the preferred value can shift across architectures. We therefore treat $\alpha$ as a model-dependent optimization hyperparameter rather than attributing its variation to model scale or parameter count.


\section{Related Work}
\label{sec:related_work}

\paragraph{Few-step diffusion distillation.}
Diffusion and flow-based generative models achieve strong synthesis quality but typically require iterative sampling, motivating extensive work on reducing the number of function evaluations at inference time. Early approaches compress multi-step teachers through progressive distillation~\citep{salimans2022progressive}, while consistency-based methods learn mappings that directly connect noisy states to clean samples or enforce consistency along probability-flow trajectories~\citep{song2023consistency,luo2023latent}. More recent methods explore trajectory-, transition-, and output-level distillation objectives to further reduce sampling steps while preserving generation quality~\citep{luo2026tdm,zhou2024score}. These approaches primarily constrain the student through teacher-generated targets or trajectories. In contrast, distribution matching methods optimize the student at the level of its induced marginal distribution, allowing the student trajectory to deviate from the teacher's original sampling path.

\paragraph{Distribution matching distillation.}
Distribution Matching Distillation (DMD)~\citep{yin2024one} introduced a
distribution-level approach to few-step generation by minimizing an approximate KL divergence between teacher and student distributions. Its gradient is estimated through the difference between a pretrained teacher score and a ``fake'' score model trained to track the evolving student distribution. DMD2~\citep{yin2024improved} substantially improves this framework by removing the original regression-data requirement, stabilizing fake-model estimation with a two-time-scale update rule, incorporating real-data adversarial supervision, and extending DMD to multi-step generation. Subsequent work has improved different aspects of the DMD framework. Diversity-Preserved DMD~\citep{wu2026diversity} separates the roles of distilled steps to mitigate diversity collapse, while Decoupled DMD~\citep{liu2025decoupled} decomposes the effects of classifier-free guidance augmentation and distribution matching and uses this decomposition to redesign the distillation schedule. Other studies investigate broader design choices surrounding distribution matching, including training recipes, teacher guidance, and data composition~\citep{wu2026qwenimageflashobjectivedesign}. Together, these works demonstrate that both the objective and the surrounding optimization procedure are critical to successful few-step distribution
matching.

\paragraph{Adaptive optimization of DMD.}
A growing line of work further studies when standard DMD supervision may become ineffective or unstable. Adaptive Matching Distillation
(AMD)~\citep{bai2026optimizing}, for example, identifies failure regions in the DMD optimization landscape and employs reward-based proxies together with adaptive correction mechanisms to steer the student away from such regions. Other approaches introduce auxiliary perceptual, adversarial, reward, or task-specific signals to improve fidelity, diversity, or training stability~\citep{jiang2025distribution,wu2026diversity}. These methods highlight the importance of adapting DMD optimization, but typically do so by augmenting the supervision, modifying individual objective components, or introducing additional mechanisms.

Our work is complementary to these directions. Rather than adding a new source of supervision or explicitly detecting failure states with an external proxy, we ask whether DMD's native teacher--fake discrepancy can itself diagnose the aggressiveness of its updates. We find that disagreement magnitude contains such information and use it as a continuous control signal for selective high-noise attenuation. DA-DMD therefore modifies how strongly existing DMD supervision is applied, rather than introducing a new target for the student.

\section{Additional Theoretical Analysis}
\label{app:theory}

This section provides the complete derivations and complementary interpretations underlying the theoretical results in Secs.~\ref{sec:dmd}--\ref{sec:theory}. We first derive the score--clean prediction identity used to express the DMD update in terms of teacher--fake clean predictions. We then analyze disagreement-aware attenuation from optimization, robustness, and gradient-estimation perspectives, followed by proximal and probabilistic interpretations of the practical weighting rule.

\subsection{Score--Clean Prediction Identity}
\label{app:score_derivation}

We first derive Eq.~\ref{eq:score_x0_identity}. Consider the Gaussian perturbation process
\begin{equation}
\label{eq:app_forward}
x_t
=
\alpha_t x_0
+
\sigma_t\epsilon,
\qquad
\epsilon\sim\mathcal N(0,I),
\end{equation}
which induces the conditional transition
\begin{equation}
\label{eq:app_transition}
q_t(x_t\mid x_0)
=
\mathcal N
\left(
x_t;\alpha_t x_0,\sigma_t^2I
\right).
\end{equation}
For a conditional clean-data distribution \(p_0(x_0\mid c)\), the noisy marginal is
\begin{equation}
\label{eq:app_marginal}
p_t(x_t\mid c)
=
\int
p_0(x_0\mid c)
q_t(x_t\mid x_0)
\,dx_0.
\end{equation}

Applying Fisher's identity gives
\begin{align}
s_t(x_t,c)
&=
\nabla_{x_t}\log p_t(x_t\mid c)
\nonumber\\
&=
\mathbb E
\left[
\nabla_{x_t}\log q_t(x_t\mid x_0)
\mid x_t,c
\right].
\label{eq:app_fisher_identity}
\end{align}
For the Gaussian transition in Eq.~\ref{eq:app_transition},
\begin{equation}
\label{eq:app_transition_score}
\nabla_{x_t}\log q_t(x_t\mid x_0)
=
-\frac{x_t-\alpha_t x_0}{\sigma_t^2}.
\end{equation}
Substituting Eq.~\ref{eq:app_transition_score} into Eq.~\ref{eq:app_fisher_identity} yields
\begin{equation}
\label{eq:app_tweedie}
s_t(x_t,c)
=
\frac{
\alpha_t\mathbb E[x_0\mid x_t,c]-x_t
}{
\sigma_t^2
},
\end{equation}
which recovers Eq.~\ref{eq:score_x0_identity}.

For an \(x_0\)-prediction model trained with squared-error denoising, the population-optimal predictor is
\begin{equation}
\label{eq:app_posterior_mean}
\hat{x}_0^\star(x_t,t,c)
=
\mathbb E[x_0\mid x_t,c].
\end{equation}
Hence, a trained model satisfies
\(\hat{x}_0(x_t,t,c)\approx\mathbb E[x_0\mid x_t,c]\), and Eq.~\ref{eq:app_tweedie} gives
\begin{equation}
\label{eq:app_score_x0}
s_t(x_t,t,c)
\approx
\frac{
\alpha_t\hat{x}_0(x_t,t,c)-x_t
}{
\sigma_t^2
}.
\end{equation}

Since the teacher and fake model are evaluated at the same noisy sample \(x_t\), the shared term \(-x_t/\sigma_t^2\) cancels:
\begin{equation}
\label{eq:app_teacher_fake_score}
s_F-s_T
\approx
\frac{\alpha_t}{\sigma_t^2}
\left(
\hat{x}_{0,F}-\hat{x}_{0,T}
\right).
\end{equation}
Thus, the clean-prediction disagreement used by DA-DMD is directly proportional, up to timestep-dependent scaling, to the native teacher--fake score discrepancy driving the DMD correction.

\subsection{One-Step Adaptive Attenuation}
\label{app:proof_attenuation}

We now provide the complete derivation of the adaptive step-control result summarized in Sec.~\ref{sec:theory}. Let \(g=g_{\mathrm{DMD}}\), and consider a scaled update
\begin{equation}
\label{eq:app_scaled_step}
\theta^+
=
\theta-\eta a g,
\qquad
a\in[0,1].
\end{equation}
Assume that the DMD objective \(\mathcal L(\theta)\) is \(\beta\)-smooth. Then
\begin{equation}
\label{eq:app_smoothness}
\mathcal L(\theta^+)
\le
\mathcal L(\theta)
-
\eta a
\langle\nabla\mathcal L(\theta),g\rangle
+
\frac{\beta\eta^2a^2}{2}
\|g\|^2.
\end{equation}

Conditioned on noise level \(t\) and disagreement \(r\), define
\begin{equation}
\label{eq:app_AB}
\begin{aligned}
A(t,r)
&=
\mathbb E
\left[
\langle\nabla\mathcal L(\theta),g\rangle
\mid t,r
\right]
>0,\\
B(t,r)
&=
\mathbb E
\left[
\|g\|^2
\mid t,r
\right]
>0.
\end{aligned}
\end{equation}
Here, \(A(t,r)\) measures useful descent alignment, while \(B(t,r)\) measures update energy. Taking the conditional expectation of Eq.~\ref{eq:app_smoothness} gives
\begin{equation}
\label{eq:app_one_step_bound}
\mathbb E
\left[
\mathcal L(\theta^+)
\mid t,r
\right]
\le
\mathcal L(\theta)
-
\eta aA(t,r)
+
\frac{\beta\eta^2a^2}{2}B(t,r).
\end{equation}

\paragraph{Optimal retained update strength.}
The \(a\)-dependent part of the bound is
\(
Q(a)
=
-\eta aA
+
\frac{\beta\eta^2a^2}{2}B
\).
Differentiating gives
\(
\partial Q/\partial a
=
-\eta A+\beta\eta^2aB
\),
whose unconstrained minimizer is \(A/(\beta\eta B)\). Enforcing \(a\in[0,1]\) therefore yields
\begin{equation}
\label{eq:app_optimal_attenuation}
a^\star(t,r)
=
\min
\left\{
1,
\frac{
A(t,r)
}{
\beta\eta B(t,r)
}
\right\}.
\end{equation}
Equivalently,
\begin{equation}
\label{eq:app_energy_alignment_relation}
\frac{B(t,r)}{A(t,r)}
\uparrow
\quad\Longrightarrow\quad
a^\star(t,r)\downarrow.
\end{equation}
Eq.~\ref{eq:app_energy_alignment_relation} formalizes the distinction between update direction and update magnitude: a DMD update may remain positively aligned with the descent direction while still being too aggressive to apply at full strength.

\paragraph{When is attenuation preferable to vanilla DMD?}
For a fixed attenuation \(a\in(0,1)\), let \(U_a\) denote the right-hand side of Eq.~\ref{eq:app_one_step_bound}, and let \(U_1\) denote the vanilla-DMD bound. Their difference is
\begin{align}
U_a-U_1
&=
\eta(1-a)A
-
\frac{\beta\eta^2}{2}
(1-a^2)B
\nonumber\\
&=
\eta(1-a)
\left[
A-
\frac{\beta\eta}{2}(1+a)B
\right].
\label{eq:app_bound_difference}
\end{align}
Since \(a\in(0,1)\), we obtain
\begin{equation}
\label{eq:app_attenuation_condition}
U_a<U_1
\quad\Longleftrightarrow\quad
\frac{B(t,r)}{A(t,r)}
>
\frac{2}{
\beta\eta(1+a)
}.
\end{equation}
Therefore, attenuation admits a strictly tighter one-step smoothness bound whenever update energy is sufficiently large relative to useful descent alignment.

\paragraph{Connection to disagreement.}
Suppose that within the active high-noise region \(\mathcal H\), the conditional energy-to-alignment ratio
\begin{equation}
\label{eq:app_ratio_assumption}
\rho(t,r)
=
\frac{B(t,r)}{A(t,r)}
\end{equation}
is nondecreasing with disagreement \(r\). Combining this assumption with Eq.~\ref{eq:app_optimal_attenuation} gives
\begin{equation}
\label{eq:app_disagreement_optimal}
r\uparrow
\quad\Longrightarrow\quad
a^\star(t,r)\downarrow.
\end{equation}
This monotonic trend is consistent with the practical DA-DMD weighting rule in Eq.~\ref{eq:da_weight}, where larger teacher--fake disagreement induces stronger attenuation. Importantly, this argument does not require high-disagreement updates to be incorrect; it only requires their update energy to grow sufficiently relative to useful descent alignment.

\subsection{Selective Energy Control and Redescending Influence}
\label{app:proof_properties}

We next establish the selective nature of DA-DMD and characterize its response to extreme disagreement.

\paragraph{Selective update-energy control.}
Ignoring the stop-gradient operator, which does not change the forward value of the multiplier, Eq.~\ref{eq:da_dmd_update} gives
\begin{equation}
\label{eq:app_da_update}
g_{\mathrm{DA-DMD}}
=
\left[
1-\chi+\chi\omega_\alpha(r)
\right]
g_{\mathrm{DMD}},
\qquad
\chi=\mathbf1[t\in\mathcal H].
\end{equation}
Since \(\chi\in\{0,1\}\),
\begin{equation}
\label{eq:app_multiplier_square}
\left[
1-\chi+\chi\omega_\alpha(r)
\right]^2
=
1-\chi
\left[
1-\omega_\alpha^2(r)
\right].
\end{equation}
Combining Eqs.~\ref{eq:app_da_update} and~\ref{eq:app_multiplier_square} yields
\begin{align}
\mathbb E
\|g_{\mathrm{DA-DMD}}\|^2
&=
\mathbb E
\|g_{\mathrm{DMD}}\|^2
\nonumber\\
&\quad-
\mathbb E
\left[
\mathbf1[t\in\mathcal H]
\left(
1-\omega_\alpha^2(r)
\right)
\|g_{\mathrm{DMD}}\|^2
\right].
\label{eq:app_energy_control}
\end{align}
Hence, the instantaneous update-energy reduction induced by DA-DMD is localized entirely to the active region. In particular,
\begin{equation}
\label{eq:app_exact_preservation}
t\notin\mathcal H
\quad\Longrightarrow\quad
g_{\mathrm{DA-DMD}}
=
g_{\mathrm{DMD}}.
\end{equation}

\paragraph{Monotonic and redescending influence.}
Recall the practical weighting rule
\begin{equation}
\label{eq:app_da_weight}
\omega_\alpha(r)
=
\frac{
\alpha e^{-r^2/2}
}{
\alpha e^{-r^2/2}+1-\alpha
}.
\end{equation}
Its derivative is
\begin{equation}
\label{eq:app_weight_derivative}
\frac{d\omega_\alpha}{dr}
=
-r\,
\omega_\alpha(r)
\left[
1-\omega_\alpha(r)
\right],
\end{equation}
which is strictly negative for \(r>0\). Thus, larger disagreement always induces stronger attenuation.

Moreover,
\begin{equation}
\label{eq:app_weight_bound}
\omega_\alpha(r)
\le
\frac{\alpha}{1-\alpha}
e^{-r^2/2}.
\end{equation}
Multiplying both sides by \(r\ge0\) gives
\begin{equation}
\label{eq:app_influence_bound}
0
\le
r\omega_\alpha(r)
\le
\frac{\alpha}{1-\alpha}
r e^{-r^2/2}.
\end{equation}
Since \(r e^{-r^2/2}\rightarrow0\) as \(r\rightarrow\infty\),
\begin{equation}
\label{eq:app_redescending}
\boxed{
\lim_{r\rightarrow\infty}
r\,\omega_\alpha(r)
=
0.
}
\end{equation}
The disagreement-dependent correction therefore exhibits redescending influence: increasingly extreme teacher--fake disagreement receives progressively diminishing effective weight instead of unbounded influence.

\paragraph{Ideal fixed-point preservation.}
At the ideal distribution-matching solution, the student matches the teacher and the fake model perfectly tracks the student. Hence \(s_F=s_T\), which implies \(g_{\mathrm{DMD}}=0\). Eq.~\ref{eq:da_dmd_update} therefore gives
\begin{equation}
\label{eq:app_fixed_point}
g_{\mathrm{DMD}}=0
\quad\Longrightarrow\quad
g_{\mathrm{DA-DMD}}=0.
\end{equation}
Thus, DA-DMD modifies the optimization dynamics while preserving the ideal distribution-matching solution. We do not require or claim that all non-ideal stationary points of DA-DMD and vanilla DMD are identical.

\subsection{Bias--Variance Interpretation}
\label{app:bias_variance}

The smoothness analysis above studies the quality of a single optimization step. A complementary view interprets disagreement-aware attenuation as conditional shrinkage of a stochastic DMD update.

Suppose
\begin{equation}
\label{eq:app_noise_model}
g
=
g^\star+\epsilon,
\qquad
\mathbb E[\epsilon\mid t,r]=0,
\end{equation}
where \(g^\star\) denotes the conditional mean update. Define the conditional stochastic variation as
\begin{equation}
\label{eq:app_variance}
V(t,r)
=
\mathbb E
\left[
\|\epsilon\|^2
\mid t,r
\right].
\end{equation}
The vanilla estimator therefore has
\(
\operatorname{MSE}_{\mathrm{DMD}}=V
\).

For a shrunk estimator \(g_a=ag\), \(a\in(0,1)\),
\(
g_a-g^\star
=
(a-1)g^\star+a\epsilon
\).
Using the conditional zero-mean assumption gives
\begin{equation}
\label{eq:app_shrinkage_mse}
\operatorname{MSE}(a)
=
(1-a)^2
\|g^\star\|^2
+
a^2V.
\end{equation}
The first term is the bias introduced by shrinkage, while the second is the remaining stochastic variation.

Comparing Eq.~\ref{eq:app_shrinkage_mse} with the vanilla estimator yields
\begin{equation}
\label{eq:app_mse_condition}
\operatorname{MSE}(a)
<
\operatorname{MSE}_{\mathrm{DMD}}
\quad\Longleftrightarrow\quad
V
>
\frac{1-a}{1+a}
\|g^\star\|^2.
\end{equation}
Minimizing Eq.~\ref{eq:app_shrinkage_mse} with respect to \(a\) gives
\begin{equation}
\label{eq:app_mse_optimal}
a^\star_{\mathrm{MSE}}
=
\frac{
\|g^\star\|^2
}{
\|g^\star\|^2+V
}.
\end{equation}
Therefore,
\begin{equation}
\label{eq:app_variance_relation}
\frac{V}{\|g^\star\|^2}
\uparrow
\quad\Longrightarrow\quad
a^\star_{\mathrm{MSE}}\downarrow.
\end{equation}
This bias--variance view complements the smoothness analysis in Sec.~\ref{app:proof_attenuation}: both favor more conservative updates when the stochastic DMD correction becomes large or uncertain relative to its useful signal.

We emphasize that this interpretation does not assume that disagreement \(r\) directly estimates \(V(t,r)\); rather, it provides an independent statistical explanation for why conditional shrinkage can improve a noisy update estimator.

\subsection{Proximal and Energy-Based Interpretations}
\label{app:proximal}

We finally provide two complementary interpretations of the practical disagreement-aware weighting rule.

\paragraph{Proximal interpretation.}
Within the active region \(t\in\mathcal H\), DA-DMD scales the original teacher-directed correction by \(\omega_\alpha(r)\). Define the adaptive proximal target
\begin{equation}
\label{eq:app_proximal_target}
\hat{x}_{0,P}
=
\left[
1-\omega_\alpha(r)
\right]
\hat{x}_{0,F}
+
\omega_\alpha(r)
\hat{x}_{0,T}.
\end{equation}
Then
\begin{equation}
\label{eq:app_proximal_difference}
\hat{x}_{0,F}-\hat{x}_{0,P}
=
\omega_\alpha(r)
\left(
\hat{x}_{0,F}-\hat{x}_{0,T}
\right).
\end{equation}
Thus, scaling the DMD correction by \(\omega_\alpha(r)\) is equivalent to replacing the full teacher-directed correction with a shorter correction toward an adaptive point between the fake and teacher predictions. As disagreement increases, \(\omega_\alpha(r)\) decreases and the effective target moves toward \(\hat{x}_{0,F}\), yielding a more conservative update. This is the geometric interpretation illustrated in Fig.~\ref{fig:method}.

\paragraph{Energy-based responsibility interpretation.}
The same weighting rule also admits an interpretation in terms of normalized energy-based responsibilities. Consider two disagreement-dependent compatibility potentials associated with the teacher-directed and fake-centered components:
\begin{equation}
\label{eq:app_energy_potentials}
\psi_T(r)
=
e^{-r^2/2},
\qquad
\psi_F(r)
=
1.
\end{equation}
Together with mixture coefficients
\(
\pi_T=\alpha
\)
and
\(
\pi_F=1-\alpha
\),
we define the normalized responsibility of the teacher-directed component as
\begin{equation}
\label{eq:app_energy_responsibility}
\rho_T(r)
=
\frac{
\pi_T\psi_T(r)
}{
\pi_T\psi_T(r)
+
\pi_F\psi_F(r)
}
=
\frac{
\alpha e^{-r^2/2}
}{
\alpha e^{-r^2/2}
+
1-\alpha
}
=
\omega_\alpha(r).
\end{equation}
Accordingly,
\(
\rho_F(r)=1-\rho_T(r)
\).
As disagreement increases, the compatibility of the teacher-directed component decreases relative to the fake-centered component, resulting in a progressively smaller teacher responsibility.

Using these responsibilities to interpolate between the two predictions gives
\begin{equation}
\label{eq:app_responsibility_target}
\rho_T(r)\hat{x}_{0,T}
+
\rho_F(r)\hat{x}_{0,F}
=
\omega_\alpha(r)\hat{x}_{0,T}
+
\left[
1-\omega_\alpha(r)
\right]
\hat{x}_{0,F},
\end{equation}
which exactly recovers the proximal target in Eq.~\ref{eq:app_proximal_target}.

Importantly, \(\psi_T\) and \(\psi_F\) are unnormalized compatibility potentials rather than probability densities over \(r\). Therefore, Eq.~\ref{eq:app_energy_responsibility} should be interpreted as an energy-based normalization of relative responsibilities, rather than as an estimate of the true teacher--fake density ratio.

\subsection{Arithmetic versus Geometric Mixtures}
\label{app:mixture_comparison}

We briefly compare two natural interpolations between the teacher and fake distributions and explain why DA-DMD adopts the arithmetic mixture.

\paragraph{Arithmetic mixture.}
Consider
\begin{equation}
p_{\alpha,t}^{\mathrm{arith}}(x_t)
=
\alpha p_{T,t}(x_t)
+
(1-\alpha)p_{F,t}(x_t).
\end{equation}
Its score is
\begin{equation}
s_{\alpha}^{\mathrm{arith}}
=
\rho_\alpha(x_t)s_T
+
\left[
1-\rho_\alpha(x_t)
\right]s_F,
\end{equation}
where
\begin{equation}
\rho_\alpha(x_t)
=
\frac{
\alpha p_{T,t}(x_t)
}{
\alpha p_{T,t}(x_t)
+
(1-\alpha)p_{F,t}(x_t)
}
=
\mathcal W_\alpha(\Delta_t(x_t)).
\end{equation}
Therefore,
\begin{equation}
s_F-s_{\alpha}^{\mathrm{arith}}
=
\rho_\alpha(x_t)
\left(
s_F-s_T
\right).
\end{equation}
Although the mixture coefficient \(\alpha\) is fixed globally, the effective score-space coefficient \(\rho_\alpha(x_t)\) depends on the sample. The arithmetic mixture therefore naturally induces sample-adaptive modulation of the original DMD correction.

\paragraph{Geometric mixture.}
Alternatively, consider the normalized geometric interpolation
\begin{equation}
p_{\alpha,t}^{\mathrm{geo}}(x_t)
=
\frac{
p_{T,t}(x_t)^\alpha
p_{F,t}(x_t)^{1-\alpha}
}{
Z_\alpha
},
\end{equation}
where \(Z_\alpha\) is the normalization constant. Since \(Z_\alpha\) does not depend on \(x_t\),
\begin{align}
s_{\alpha}^{\mathrm{geo}}
&=
\nabla_{x_t}
\log p_{\alpha,t}^{\mathrm{geo}}(x_t)
\nonumber\\
&=
\alpha s_T
+
(1-\alpha)s_F.
\end{align}
Hence,
\begin{equation}
s_F-s_{\alpha}^{\mathrm{geo}}
=
\alpha
\left(
s_F-s_T
\right).
\end{equation}
The geometric mixture therefore preserves the DMD correction direction but only rescales it by the constant factor \(\alpha\), independent of the sample.

\paragraph{Implication for DA-DMD.}
Both constructions produce intermediate distributions between the teacher and fake models, but they differ critically in score space. The geometric mixture yields uniform attenuation, whereas the arithmetic mixture converts a fixed distribution-level mixture coefficient into a sample-dependent score-space responsibility. Since our goal is to adapt the update strength according to sample-level training signals while preserving the original DMD correction direction, the arithmetic mixture provides the more appropriate structural template.

We emphasize that this comparison motivates the \emph{form} of sample-adaptive weighting. The practical DA-DMD weight further replaces the unavailable density-ratio-dependent responsibility with the disagreement-based control rule introduced in Sec.~\ref{sec:da_dmd}.

\section{Additional Experiments}
\label{app:additional_experiments}

We provide additional experiments complementing the main-paper evaluation, including controlled analyses of DMD optimization, complete quantitative results on additional benchmarks, and qualitative comparisons across all evaluated backbones.

\begin{figure*}[h]
\begin{center}
\includegraphics[width=1.\linewidth]{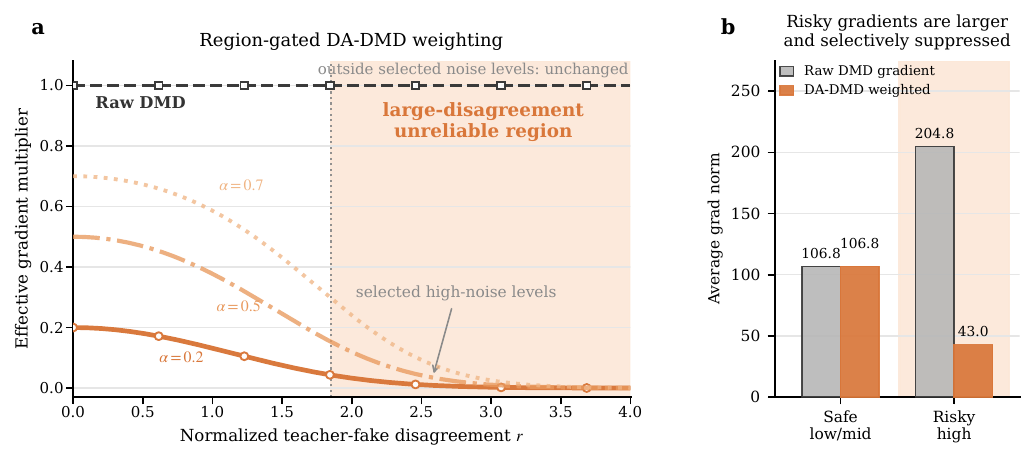}
\vspace{-0.3in}
\caption{
\textbf{Selective disagreement-aware weighting in DA-DMD.}
(a) Outside the predefined high-noise region $\mathcal H$, DA-DMD exactly preserves the vanilla DMD update. Within $\mathcal H$, the retained update strength decreases continuously with normalized teacher--fake disagreement $r$, with $\alpha$ controlling its maximum value.
(b) The resulting weighting leaves low-/mid-noise updates unchanged while substantially attenuating the high-energy updates concentrated in the high-noise regime.
}
\label{fig:motivation_weight}
\end{center}
\end{figure*}

\subsection{Diagnostic Analysis}
\label{app:diagnostic}

\begin{figure*}[h]
\begin{center}
\includegraphics[width=1.\linewidth]{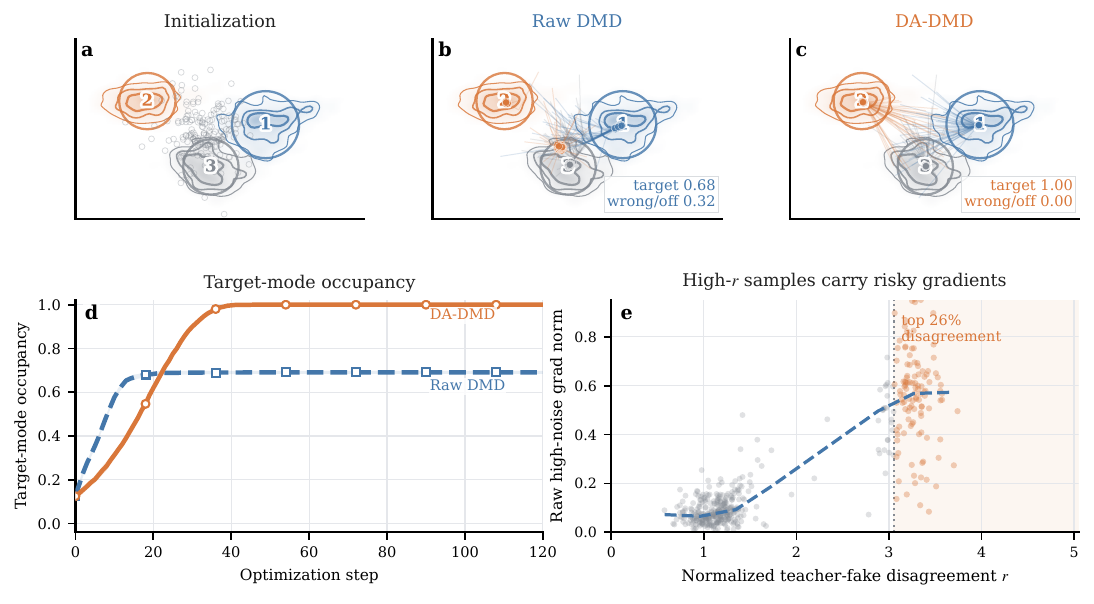}
\vspace{-0.3in}
\caption{\textbf{Controlled analysis of disagreement-aware DMD optimization on CIFAR-10.}
Starting from the same initialization, raw DMD can be dominated by high-disagreement samples with large update magnitudes, leading to an imbalanced optimization trajectory. DA-DMD selectively attenuates these updates while retaining lower-disagreement corrections, resulting in improved target-mode matching. The bottom panels visualize the evolution of target-mode occupancy and the concentration of large raw DMD updates among high-disagreement samples.
}
\label{fig:motivation_cifar10}
\end{center}
\end{figure*}

\begin{figure*}[h]
\begin{center}
\includegraphics[width=1.\linewidth]{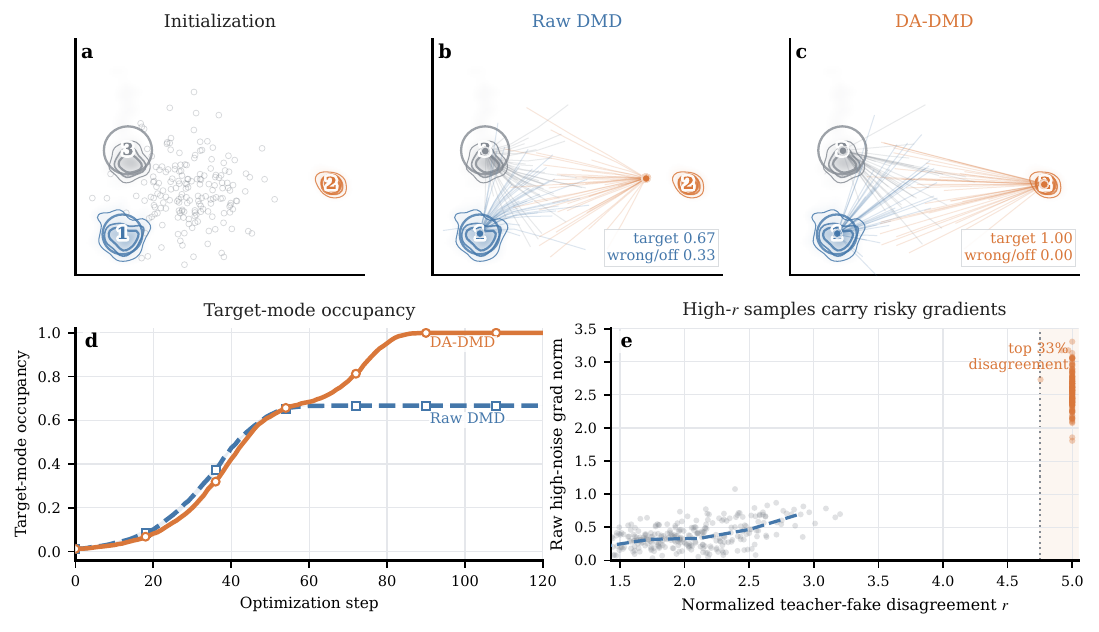}
\caption{\textbf{Controlled analysis of disagreement-aware DMD optimization on DIGITS.}
Starting from the same initialization, raw DMD can be dominated by high-disagreement samples with large update magnitudes, leading to an imbalanced optimization trajectory. DA-DMD selectively attenuates these updates while retaining lower-disagreement corrections, resulting in improved target-mode matching. The bottom panels visualize the evolution of target-mode occupancy and the concentration of large raw DMD updates among high-disagreement samples.
}
\label{fig:motivation_digits}
\end{center}
\end{figure*}

\begin{figure*}[h]
\begin{center}
\includegraphics[width=1.\linewidth]{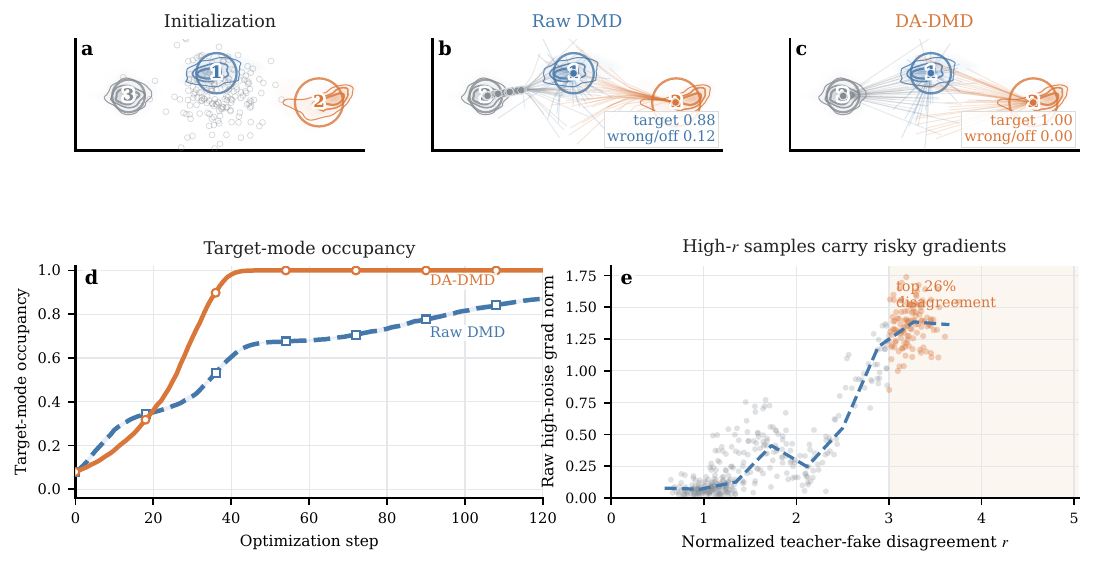}
\caption{\textbf{Controlled analysis of disagreement-aware DMD optimization on CIFAR-100.}
Starting from the same initialization, raw DMD can be dominated by high-disagreement samples with large update magnitudes, leading to an imbalanced optimization trajectory. DA-DMD selectively attenuates these updates while retaining lower-disagreement corrections, resulting in improved target-mode matching. The bottom panels visualize the evolution of target-mode occupancy and the concentration of large raw DMD updates among high-disagreement samples.
}
\label{fig:motivation_cifar100}
\end{center}
\end{figure*}

We first provide controlled analyses to further understand the relationship between teacher--fake disagreement and DMD optimization. Figs.~\ref{fig:motivation_cifar10}, \ref{fig:motivation_digits}, and~\ref{fig:motivation_cifar100} show results on DIGITS, CIFAR-10, and CIFAR-100, respectively. Starting from the same initialization, high-disagreement samples consistently exhibit larger raw DMD updates and can disproportionately influence the optimization trajectory, leading to biased target-mode occupancy. DA-DMD attenuates these aggressive corrections and recovers the target distribution more faithfully across all three settings.

Fig.~\ref{fig:motivation_weight} further illustrates the selective weighting mechanism of DA-DMD. Updates outside the predefined high-noise region remain exactly unchanged, whereas disagreement continuously controls the retained update strength within the active region. Together with the large-scale SD3.5-M analysis in Fig.~\ref{fig:training_gradient}, these results consistently support the central design principle of DA-DMD: noise level determines \emph{where} intervention is useful, while teacher--fake disagreement determines \emph{how strongly} the update should be moderated.

\subsection{Additional Quantitative Results}
\label{app:quantitative}

\paragraph{PartiPrompts.}
Tab.~\ref{tab:parti_main} reports the complete results on PartiPrompts. DA-DMD achieves higher average pairwise win rates than DMD2 across all five backbones, demonstrating that its improvements extend beyond DrawBench to a substantially larger and more diverse prompt set. The gains are particularly pronounced on SD3.5-M and SD3.5-L, where DA-DMD reaches average win rates of $61.2\%$ and $61.4\%$, respectively.

\paragraph{GenEval2.}
Tabs.~\ref{tab:geneval_all_models} and~\ref{tab:geneval_atoms} evaluate compositional generation from complementary perspectives. DA-DMD improves the overall GenEval2 score across the evaluated backbones, with the largest gain of $+6.81$ on SD3.5-L. Improvements become especially pronounced on more compositionally demanding prompts: on SD3.5-L, DA-DMD improves the 7- and 8-atom scores by $+13.04$ and $+9.56$, respectively. These results suggest that disagreement-aware update control is particularly beneficial when multiple visual constraints must be satisfied simultaneously.

\paragraph{TIIF-Bench.}
Tabs.~\ref{tab:tiif_basic} and~\ref{tab:tiif_advanced} provide a detailed evaluation of fine-grained instruction following. DA-DMD improves the overall TIIF-Bench score across all five evaluated backbones, with the largest gain of $+2.44$ on SD3.5-L. Category-level performance varies across individual skills, but the aggregate results show broad improvements on both short- and long-form instructions, including advanced compositional settings.

\subsection{Additional Qualitative Results}
\label{app:qualitative}

\begin{figure*}[h]
\begin{center}
\includegraphics[width=1.\linewidth]{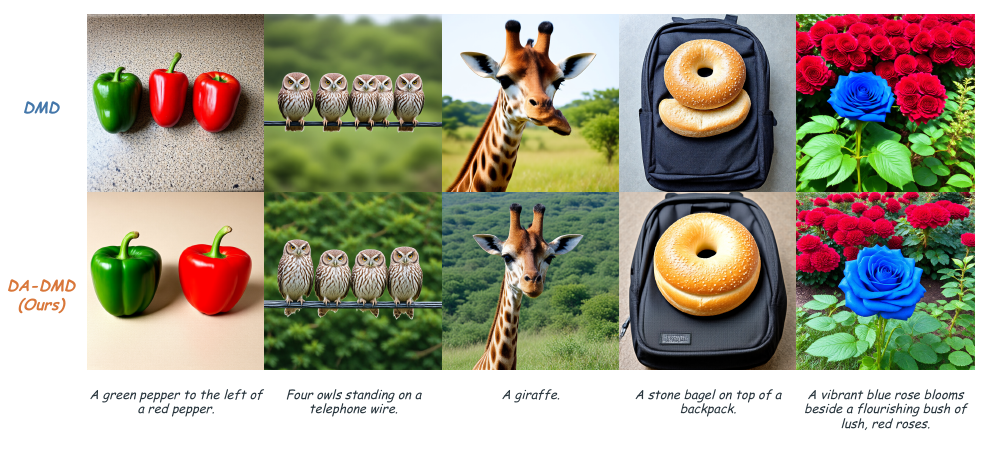}
\caption{
\textbf{Qualitative comparison on SD3-M.}
DA-DMD improves prompt adherence over DMD2 across diverse compositional challenges, including object counting, spatial placement, relative relations, color control, and structured multi-object generation.
}
\label{fig:app:sd3m}
\end{center}
\end{figure*}

\begin{figure*}[h]
\begin{center}
\includegraphics[width=1.\linewidth]{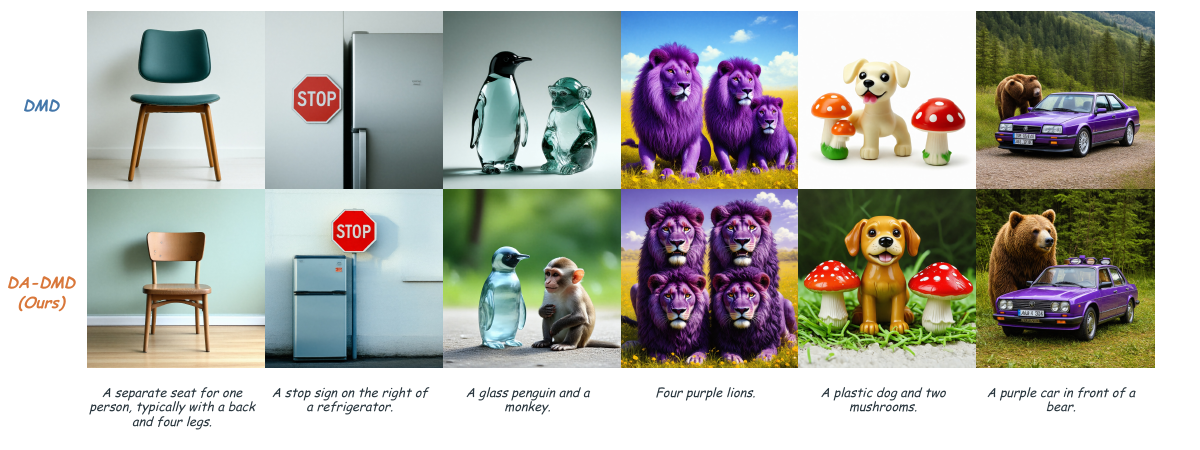}
\caption{
\textbf{Qualitative comparison on SD3.5-M.}
DA-DMD improves prompt adherence over DMD2 across diverse compositional challenges, including object counting, spatial placement, relative relations, color control, and structured multi-object generation.
}
\label{fig:app:sd35m}
\end{center}
\end{figure*}

\begin{figure*}[h]
\begin{center}
\includegraphics[width=1.\linewidth]{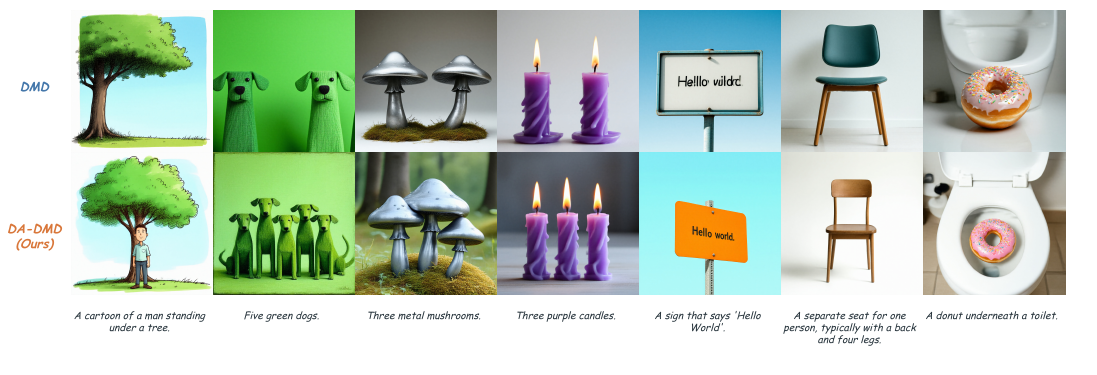}
\caption{
\textbf{Qualitative comparison on SD3.5-L.}
DA-DMD improves prompt adherence over DMD2 across diverse compositional challenges, including object counting, spatial placement, relative relations, color control, and structured multi-object generation.
}
\label{fig:app:sd35l}
\end{center}
\end{figure*}

\begin{figure*}[h]
\begin{center}
\includegraphics[width=1.\linewidth]{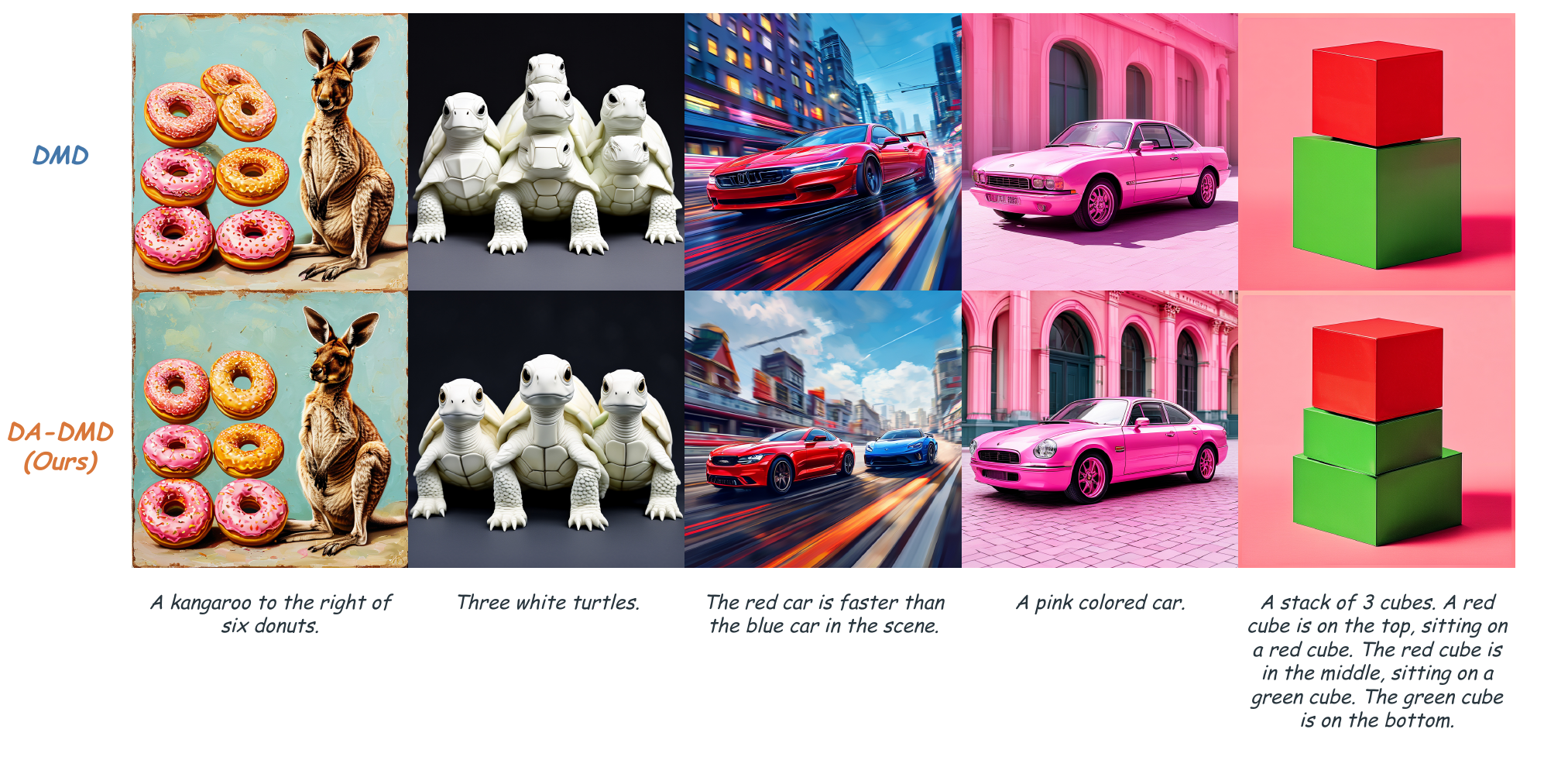}
\caption{
\textbf{Qualitative comparison on SANA.}
DA-DMD improves prompt adherence over DMD2 across diverse compositional challenges, including object counting, spatial placement, relative relations, color control, and structured multi-object generation.
}
\label{fig:app:sana}
\end{center}
\end{figure*}

\begin{figure*}[h]
\begin{center}
\includegraphics[width=1.\linewidth]{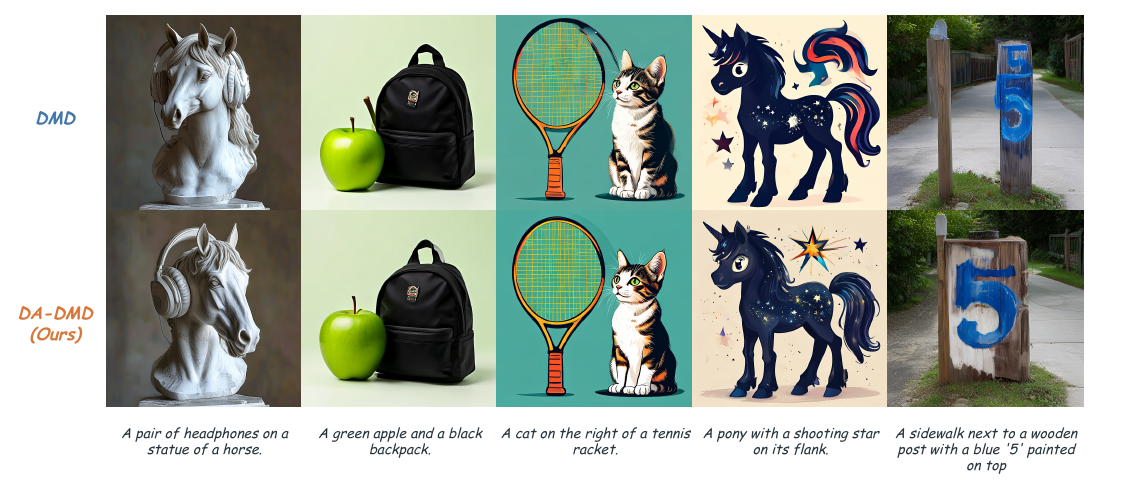}
\caption{
\textbf{Qualitative comparison on SANA-1.5.}
DA-DMD improves prompt adherence over DMD2 across diverse compositional challenges, including object counting, spatial placement, relative relations, color control, and structured multi-object generation.
}
\label{fig:app:sana15}
\end{center}
\end{figure*}

We further compare DA-DMD with DMD2 across all evaluated model families. Figs.~\ref{fig:app:sd3m}, \ref{fig:app:sd35m}, \ref{fig:app:sd35l}, \ref{fig:app:sana}, and~\ref{fig:app:sana15} present additional examples on SD3-M, SD3.5-M, SD3.5-L, SANA, and SANA-1.5, respectively. Across diverse prompts involving counting, spatial relations, relative attributes, color binding, text rendering, and structured multi-object composition, DA-DMD more faithfully satisfies the specified visual constraints while maintaining comparable perceptual quality. These examples complement the quantitative results and show that the improvements produced by disagreement-aware update control are also reflected in visually observable prompt adherence.

\begin{table*}[ht]
\centering
\caption{
\textbf{Quantitative comparison on PartiPrompts.}
DA-DMD yields broad improvements over the corresponding DMD2 baselines across different model families. 
}
\label{tab:parti_main}
\small
\setlength{\tabcolsep}{4.7pt}
\renewcommand{\arraystretch}{1.14}
\resizebox{0.95\textwidth}{!}{
\begin{tabular}{llcccccc}
\toprule
\textbf{Backbone} & \textbf{Method} & \textbf{PickScore}$\uparrow$ & \textbf{ImageReward}$\uparrow$ & \textbf{HPSv2}$\uparrow$ & \textbf{AES}$\uparrow$ & \textbf{VQ-R1}$\uparrow$ & \textbf{Avg. WR}$\uparrow$ \\
\midrule
\multirow{2}{*}[-0.9ex]{SANA} & DMD2 & 22.74 & \underline{117.28} & \textbf{30.57} & 5.940 & 4.617 & - \\
\cmidrule(lr){2-7}
 & \textcolor{TrustOrange}{\textbf{DA-DMD (Ours)}} & \ourscell{\textbf{22.77}}{+0.03} & \ourscell{\textbf{119.58}}{+2.30} & \ourscell{\textbf{30.57}}{+0.00} & \ourscell{\textbf{5.956}}{+0.016} & \ourscell{\textbf{4.638}}{+0.020} & \textbf{53.6\%} \\
\midrule
\multirow{2}{*}[-0.9ex]{SANA-1.5} & DMD2 & 22.85 & \textbf{122.79} & \underline{30.31} & 6.022 & 4.553 & - \\
\cmidrule(lr){2-7}
 & \textcolor{TrustOrange}{\textbf{DA-DMD (Ours)}} & \ourscell{\textbf{22.90}}{+0.05}  & \ourscell{\underline{122.29}}{-0.50} & \ourscell{\textbf{30.44}}{+0.13} & \ourscell{\textbf{6.050}}{+0.029} & \ourscell{\textbf{4.575}}{+0.023} & \textbf{54.6\%} \\
\midrule
\multirow{6}{*}{SD3-M} & DMD2 & 22.39 & \underline{116.74} & \underline{30.44} & 5.697 & 4.571 & - \\
 & Hyper-SD3 & 22.24 & 92.07 & 27.87 & 5.530 & 4.418 & 30.0\% \\
 & SiD-DiT & \underline{22.56} & 109.77 & 28.98 & 5.499 & 4.617 & 38.8\% \\
 & SiD-Adv & 22.53 & 104.79 & 29.86 & \underline{5.716} & 4.583 & 44.4\% \\
 & TDM & 22.54 & \textbf{117.95} & 30.38 & 5.647 & \underline{4.648} & \underline{50.2\%} \\
\cmidrule(lr){2-7}
 & \textbf{\textcolor{TrustOrange}{DA-DMD (Ours)}} & \ourscell{\textbf{22.57}}{+0.17} & \ourscell{113.96}{-2.78} & \ourscell{\textbf{30.84}}{+0.40} & \ourscell{\textbf{5.726}}{+0.030} & \ourscell{\textbf{4.678}}{+0.107} & \textbf{57.2\%} \\
\midrule
\multirow{6}{*}{SD3.5-M} & DMD2 & 22.69 & \underline{119.55} & 29.56 & 5.543 & 4.611 & - \\
 & SiD-DiT & \underline{22.71} & 114.97 & 29.87 & 5.661 & 4.661 & 49.0\% \\
 & SiD-Adv & 22.67 & 114.72 & 29.45 & \textbf{5.816} & 4.553 & 48.4\% \\
 & DP-DMD & 22.48 & 114.12 & 28.83 & 5.486 & 4.560 & 36.6\% \\
 & SenseFlow & 22.50 & 104.05 & \underline{30.18} & \underline{5.714} & \underline{4.674} & \underline{56.2\%} \\
\cmidrule(lr){2-7}
 & \textbf{\textcolor{TrustOrange}{DA-DMD (Ours)}} & \ourscell{\textbf{22.75}}{+0.06} & \ourscell{\textbf{123.73}}{+4.18} & \ourscell{\textbf{30.33}}{+0.77} & \ourscell{5.659}{+0.116} & \ourscell{\textbf{4.692}}{+0.081} & \textbf{61.2\%} \\
\midrule
\multirow{6}{*}{SD3.5-L} & DMD2 & 22.69 & 116.32 & 29.14 & 5.595 & 4.591 & - \\
 & SiD-DiT & 22.81 & 112.08 & 29.95 & 5.683 & 4.621 & 54.0\% \\
 & SiD-Adv & \underline{22.98} & 118.46 & 30.01 & \textbf{5.849} & \underline{4.628} & \underline{61.2\%} \\
 & SD3.5-L Turbo & 22.85 & 110.84 & 29.14 & \underline{5.755} & 4.616 & 52.4\% \\
 & SenseFlow & 22.89 & \underline{119.52} & \underline{30.48} & 5.639 & \textbf{4.671} & 60.8\% \\
\cmidrule(lr){2-7}
 & \textbf{\textcolor{TrustOrange}{DA-DMD (Ours)}} & \ourscell{\textbf{23.10}}{+0.41} & \ourscell{\textbf{121.78}}{+5.46} & \ourscell{\textbf{30.53}}{+1.39} & \ourscell{5.694}{+0.099} & \ourscell{4.588}{-0.003} & \textbf{61.4\%} \\
\bottomrule
\end{tabular}
}
\end{table*}



\begin{table}[t]
\centering
\caption{\textbf{GenEval2 results across backbones.}
$\Delta$ denotes the absolute improvement over DMD2.
WR is the pairwise win rate of DA-DMD against DMD2.}
\label{tab:geneval_all_models}
\small
\setlength{\tabcolsep}{8pt}
\renewcommand{\arraystretch}{1.15}
\begin{tabular}{@{}lrrrr@{}}
\toprule
\multirow{2}{*}{Backbone}
& \multicolumn{3}{c}{Overall $\uparrow$}
& \multirow{2}{*}{WR (\%) $\uparrow$} \\
\cmidrule(lr){2-4}
& DMD2
& \textcolor{TrustOrange}{DA-DMD}
& $\Delta$
& \\
\midrule
SANA
& 14.68 & \textbf{15.89}
& \textcolor{TrustOrange}{+1.21}
& 55.0 \\
SD3.5-M
& 20.17 & \textbf{21.74}
& \textcolor{TrustOrange}{+1.57}
& 55.4 \\
SD3.5-L
& 15.93 & \textbf{22.74}
& \textcolor{TrustOrange}{+6.81}
& 61.9 \\
\bottomrule
\end{tabular}
\end{table}

\begin{table*}[h]
\centering
\caption{\textbf{GenEval2 results across prompt complexity.} DA-DMD provides broad gains across compositional complexity levels, with particularly strong improvements on challenging high-atom prompts, e.g., $+13.04$ and $+9.56$ on 7- and 8-atom prompts with SD3.5-L. Colored values denote absolute changes over DMD2.}
\label{tab:geneval_atoms}
\scriptsize
\setlength{\tabcolsep}{3.8pt}
\renewcommand{\arraystretch}{1.13}
\begin{tabular}{llcccccccc}
\toprule
\textbf{Backbone} & \textbf{Method} & \textbf{3}$\uparrow$ & \textbf{4}$\uparrow$ & \textbf{5}$\uparrow$ & \textbf{6}$\uparrow$ & \textbf{7}$\uparrow$ & \textbf{8}$\uparrow$ & \textbf{9}$\uparrow$ & \textbf{10}$\uparrow$ \\
\midrule
\multirow{3}{*}{SANA} & DMD2 & 42.03 & \textbf{32.25} & \textbf{14.39} & 11.27 & 6.48 & 5.53 & 3.22 & 2.23 \\
\cmidrule(lr){2-10}
 & \textcolor{TrustOrange}{\textbf{DA-DMD}} & \makecell{\textbf{44.68}\\[-2pt]\dmdgain{+2.65}} & \makecell{32.11\\[-2pt]\dmdgain{-0.14}} & \makecell{13.89\\[-2pt]\dmdgain{-0.50}} & \makecell{\textbf{13.08}\\[-2pt]\dmdgain{+1.81}} & \makecell{\textbf{10.41}\\[-2pt]\dmdgain{+3.93}} & \makecell{\textbf{7.08}\\[-2pt]\dmdgain{+1.55}} & \makecell{\textbf{4.22}\\[-2pt]\dmdgain{+1.00}} & \makecell{\textbf{3.37}\\[-2pt]\dmdgain{+1.14}} \\
\midrule
\multirow{3}{*}{SANA-1.5} & DMD2 & 53.69 & 43.07 & 20.28 & \textbf{18.02} & \textbf{6.95} & \textbf{9.27} & 5.63 & 3.43 \\
\cmidrule(lr){2-10}
 & \textcolor{TrustOrange}{\textbf{DA-DMD}} & \makecell{\textbf{54.66}\\[-2pt]\dmdgain{+0.97}} & \makecell{\textbf{44.20}\\[-2pt]\dmdgain{+1.13}} & \makecell{\textbf{22.62}\\[-2pt]\dmdgain{+2.34}} & \makecell{17.97\\[-2pt]\dmdgain{-0.05}} & \makecell{6.59\\[-2pt]\dmdgain{-0.36}} & \makecell{8.84\\[-2pt]\dmdgain{-0.43}} & \makecell{\textbf{6.73}\\[-2pt]\dmdgain{+1.10}} & \makecell{\textbf{3.98}\\[-2pt]\dmdgain{+0.55}} \\
\midrule
\multirow{2}{*}{SD3.5-M} & DMD2 & 50.01 & \textbf{39.16} & 26.65 & 18.27 & 8.44 & 7.75 & 7.90 & 3.14 \\
\cmidrule(lr){2-10}
 & \textcolor{TrustOrange}{\textbf{DA-DMD}} & \makecell{\textbf{53.47}\\[-2pt]\dmdgain{+3.46}} & \makecell{36.63\\[-2pt]\dmdgain{-2.53}} & \makecell{\textbf{27.68}\\[-2pt]\dmdgain{+1.03}} & \makecell{\textbf{19.05}\\[-2pt]\dmdgain{+0.78}} & \makecell{\textbf{15.15}\\[-2pt]\dmdgain{+6.71}} & \makecell{\textbf{9.02}\\[-2pt]\dmdgain{+1.27}} & \makecell{\textbf{8.58}\\[-2pt]\dmdgain{+0.68}} & \makecell{\textbf{4.32}\\[-2pt]\dmdgain{+1.18}} \\
\midrule
\cmidrule(lr){2-10}
\multirow{2}{*}{SD3.5-L} & DMD2 & 39.87 & 31.37 & 20.27 & 14.98 & 6.92 & 5.83 & 5.16 & 3.02 \\
\cmidrule(lr){2-10}
 & \textcolor{TrustOrange}{\textbf{DA-DMD}} & \makecell{\textbf{45.83}\\[-2pt]\dmdgain{+5.96}} & \makecell{\textbf{42.11}\\[-2pt]\dmdgain{+10.74}} & \makecell{\textbf{26.81}\\[-2pt]\dmdgain{+6.54}} & \makecell{\textbf{20.10}\\[-2pt]\dmdgain{+5.12}} & \makecell{\textbf{19.96}\\[-2pt]\dmdgain{+13.04}} & \makecell{\textbf{15.39}\\[-2pt]\dmdgain{+9.56}} & \makecell{\textbf{7.81}\\[-2pt]\dmdgain{+2.65}} & \makecell{\textbf{3.89}\\[-2pt]\dmdgain{+0.87}} \\
\bottomrule
\end{tabular}
\end{table*}


\begin{table*}[t]
\centering
\caption{\textbf{TIIF-Bench overall and basic-category results.} DA-DMD improves the overall instruction-following score across all five evaluated backbones, with the largest gain of $+2.44$ on SD3.5-L, while showing broadly positive improvements across short- and long-form categories.}
\label{tab:tiif_basic}
\scriptsize
\setlength{\tabcolsep}{3.8pt}
\renewcommand{\arraystretch}{1.14}
\resizebox{\textwidth}{!}{%
\begin{tabular}{llccccccccccc}
\toprule
\textbf{Backbone} & \textbf{Method} & \textbf{Overall}$\uparrow$ & \textbf{Overall S}$\uparrow$ & \textbf{Overall L}$\uparrow$ & \textbf{Basic S}$\uparrow$ & \textbf{Basic L}$\uparrow$ & \textbf{Attr. S}$\uparrow$ & \textbf{Attr. L}$\uparrow$ & \textbf{Rela. S}$\uparrow$ & \textbf{Rela. L}$\uparrow$ & \textbf{Reason S}$\uparrow$ & \textbf{Reason L}$\uparrow$ \\
\midrule
\multirow{2}{*}{SANA} & DMD2 & 60.72 & 60.1 & 61.3 & \textbf{75.8} & \textbf{77.3} & \textbf{79.3} & 79.6 & 84.1 & \textbf{84.6} & \textbf{63.4} & \textbf{67.6} \\
\cmidrule(lr){2-13}
 & \textcolor{TrustOrange}{\textbf{DA-DMD}} & \makecell{\textbf{61.12}\\[-2pt]\dmdgain{+0.40}} & \makecell{\textbf{60.4}\\[-2pt]\dmdgain{+0.3}} & \makecell{\textbf{61.8}\\[-2pt]\dmdgain{+0.5}} & \makecell{75.7\\[-2pt]\dmdgain{-0.1}} & \makecell{77.1\\[-2pt]\dmdgain{-0.2}} & \makecell{79.1\\[-2pt]\dmdgain{-0.2}} & \makecell{\textbf{79.8}\\[-2pt]\dmdgain{+0.2}} & \makecell{\textbf{84.2}\\[-2pt]\dmdgain{+0.1}} & \makecell{84.3\\[-2pt]\dmdgain{-0.3}} & \makecell{\textbf{63.4}\\[-2pt]\dmdgain{+0.0}} & \makecell{66.9\\[-2pt]\dmdgain{-0.7}} \\
\midrule
\multirow{2}{*}{SANA-1.5} & DMD2 & 62.47 & 63.2 & 61.8 & 77.1 & 76.4 & \textbf{81.2} & \textbf{79.2} & 84.5 & 83.9 & 65.1 & 65.7 \\
\cmidrule(lr){2-13}
 & \textcolor{TrustOrange}{\textbf{DA-DMD}} & \makecell{\textbf{62.84}\\[-2pt]\dmdgain{+0.37}} & \makecell{\textbf{63.5}\\[-2pt]\dmdgain{+0.3}} & \makecell{\textbf{62.2}\\[-2pt]\dmdgain{+0.4}} & \makecell{\textbf{77.5}\\[-2pt]\dmdgain{+0.4}} & \makecell{\textbf{76.6}\\[-2pt]\dmdgain{+0.2}} & \makecell{80.8\\[-2pt]\dmdgain{-0.4}} & \makecell{79.1\\[-2pt]\dmdgain{-0.1}} & \makecell{\textbf{84.8}\\[-2pt]\dmdgain{+0.3}} & \makecell{\textbf{84.7}\\[-2pt]\dmdgain{+0.8}} & \makecell{\textbf{66.3}\\[-2pt]\dmdgain{+1.2}} & \makecell{\textbf{66.1}\\[-2pt]\dmdgain{+0.4}} \\
\midrule
\multirow{2}{*}{SD3-M} & DMD2 & 63.33 & 63.8 & 62.8 & 77.8 & 77.8 & 83.5 & 82.5 & \textbf{82.3} & \textbf{83.1} & 66.4 & \textbf{66.9} \\
\cmidrule(lr){2-13}
 & \textcolor{TrustOrange}{\textbf{DA-DMD}} & \makecell{\textbf{64.23}\\[-2pt]\dmdgain{+0.90}} & \makecell{\textbf{64.5}\\[-2pt]\dmdgain{+0.7}} & \makecell{\textbf{63.9}\\[-2pt]\dmdgain{+1.1}} & \makecell{\textbf{78.0}\\[-2pt]\dmdgain{+0.2}} & \makecell{\textbf{78.4}\\[-2pt]\dmdgain{+0.6}} & \makecell{\textbf{83.9}\\[-2pt]\dmdgain{+0.4}} & \makecell{\textbf{85.1}\\[-2pt]\dmdgain{+2.6}} & \makecell{82.0\\[-2pt]\dmdgain{-0.3}} & \makecell{81.9\\[-2pt]\dmdgain{-1.2}} & \makecell{\textbf{66.9}\\[-2pt]\dmdgain{+0.5}} & \makecell{66.7\\[-2pt]\dmdgain{-0.2}} \\
\midrule
\multirow{2}{*}{SD3.5-M} & DMD2 & 66.77 & \textbf{67.9} & 65.6 & \textbf{79.5} & 79.3 & \textbf{85.9} & 84.0 & 84.4 & \textbf{85.5} & \textbf{66.8} & 67.6 \\
\cmidrule(lr){2-13}
 & \textcolor{TrustOrange}{\textbf{DA-DMD}} & \makecell{\textbf{66.98}\\[-2pt]\dmdgain{+0.21}} & \makecell{67.5\\[-2pt]\dmdgain{-0.4}} & \makecell{\textbf{66.5}\\[-2pt]\dmdgain{+0.9}} & \makecell{78.7\\[-2pt]\dmdgain{-0.8}} & \makecell{\textbf{79.8}\\[-2pt]\dmdgain{+0.5}} & \makecell{84.3\\[-2pt]\dmdgain{-1.6}} & \makecell{\textbf{85.2}\\[-2pt]\dmdgain{+1.2}} & \makecell{\textbf{85.2}\\[-2pt]\dmdgain{+0.8}} & \makecell{85.4\\[-2pt]\dmdgain{-0.1}} & \makecell{65.7\\[-2pt]\dmdgain{-1.1}} & \makecell{\textbf{68.0}\\[-2pt]\dmdgain{+0.4}} \\
\midrule
\multirow{2}{*}{SD3.5-L} & DMD2 & 65.82 & 68.4 & 63.3 & 77.8 & 75.6 & 84.6 & 80.2 & 83.3 & 81.4 & 64.2 & 64.3 \\
\cmidrule(lr){2-13}
 & \textcolor{TrustOrange}{\textbf{DA-DMD}} & \makecell{\textbf{68.26}\\[-2pt]\dmdgain{+2.44}} & \makecell{\textbf{70.4}\\[-2pt]\dmdgain{+2.0}} & \makecell{\textbf{66.1}\\[-2pt]\dmdgain{+2.8}} & \makecell{\textbf{79.7}\\[-2pt]\dmdgain{+1.9}} & \makecell{\textbf{77.8}\\[-2pt]\dmdgain{+2.2}} & \makecell{\textbf{86.5}\\[-2pt]\dmdgain{+1.9}} & \makecell{\textbf{82.2}\\[-2pt]\dmdgain{+2.0}} & \makecell{\textbf{84.4}\\[-2pt]\dmdgain{+1.1}} & \makecell{\textbf{83.1}\\[-2pt]\dmdgain{+1.7}} & \makecell{\textbf{66.7}\\[-2pt]\dmdgain{+2.5}} & \makecell{\textbf{67.1}\\[-2pt]\dmdgain{+2.8}} \\
\bottomrule
\end{tabular}
}%
\end{table*}

\begin{table*}[t]
\centering
\caption{\textbf{TIIF-Bench advanced compositional results.} DA-DMD generally improves complex compositional instruction following across model backbones, with particularly strong gains on SD3.5-L, including $+2.4$ and $+3.3$ on advanced short- and long-form prompts. Colored values denote absolute changes over DMD2.}
\label{tab:tiif_advanced}
\scriptsize
\setlength{\tabcolsep}{3.8pt}
\renewcommand{\arraystretch}{1.14}
\resizebox{\textwidth}{!}{%
\begin{tabular}{llcccccccc}
\toprule
\textbf{Backbone} & \textbf{Method} & \textbf{Adv. S}$\uparrow$ & \textbf{Adv. L}$\uparrow$ & \textbf{Attr.+Rela. S}$\uparrow$ & \textbf{Attr.+Rela. L}$\uparrow$ & \textbf{Attr.+Rea. S}$\uparrow$ & \textbf{Attr.+Rea. L}$\uparrow$ & \textbf{Rela.+Rea. S}$\uparrow$ & \textbf{Rela.+Rea. L}$\uparrow$ \\
\midrule
\multirow{2}{*}{SANA} & DMD2 & 51.8 & 52.8 & 78.0 & 76.3 & 63.3 & \textbf{66.9} & 66.0 & 68.7 \\
\cmidrule(lr){2-10}
 & \textcolor{TrustOrange}{\textbf{DA-DMD}} & \makecell{\textbf{52.3}\\[-2pt]\dmdgain{+0.5}} & \makecell{\textbf{53.8}\\[-2pt]\dmdgain{+1.0}} & \makecell{\textbf{78.2}\\[-2pt]\dmdgain{+0.2}} & \makecell{\textbf{78.2}\\[-2pt]\dmdgain{+1.9}} & \makecell{\textbf{63.5}\\[-2pt]\dmdgain{+0.2}} & \makecell{\textbf{66.9}\\[-2pt]\dmdgain{+0.0}} & \makecell{\textbf{67.0}\\[-2pt]\dmdgain{+1.0}} & \makecell{\textbf{71.0}\\[-2pt]\dmdgain{+2.3}} \\
\midrule
\multirow{2}{*}{SANA-1.5} & DMD2 & 56.1 & 54.0 & 80.4 & 76.7 & 68.7 & 65.9 & \textbf{68.8} & \textbf{69.4} \\
\cmidrule(lr){2-10}
 & \textcolor{TrustOrange}{\textbf{DA-DMD}} & \makecell{\textbf{56.3}\\[-2pt]\dmdgain{+0.2}} & \makecell{\textbf{54.7}\\[-2pt]\dmdgain{+0.7}} & \makecell{\textbf{80.6}\\[-2pt]\dmdgain{+0.2}} & \makecell{\textbf{76.8}\\[-2pt]\dmdgain{+0.1}} & \makecell{\textbf{69.8}\\[-2pt]\dmdgain{+1.1}} & \makecell{\textbf{67.2}\\[-2pt]\dmdgain{+1.3}} & \makecell{68.4\\[-2pt]\dmdgain{-0.4}} & \makecell{68.6\\[-2pt]\dmdgain{-0.8}} \\
\midrule
\multirow{2}{*}{SD3-M} & DMD2 & 57.1 & 55.4 & 78.0 & 78.7 & 63.9 & 65.7 & 63.2 & 66.9 \\
\cmidrule(lr){2-10}
 & \textcolor{TrustOrange}{\textbf{DA-DMD}} & \makecell{\textbf{58.2}\\[-2pt]\dmdgain{+1.1}} & \makecell{\textbf{57.0}\\[-2pt]\dmdgain{+1.6}} & \makecell{\textbf{78.8}\\[-2pt]\dmdgain{+0.8}} & \makecell{\textbf{79.2}\\[-2pt]\dmdgain{+0.5}} & \makecell{\textbf{65.5}\\[-2pt]\dmdgain{+1.6}} & \makecell{\textbf{66.9}\\[-2pt]\dmdgain{+1.2}} & \makecell{\textbf{65.2}\\[-2pt]\dmdgain{+2.0}} & \makecell{\textbf{68.6}\\[-2pt]\dmdgain{+1.7}} \\
\midrule
\multirow{2}{*}{SD3.5-M} & DMD2 & \textbf{62.5} & 58.9 & \textbf{80.9} & 78.7 & \textbf{68.2} & \textbf{67.4} & 66.4 & 67.6 \\
\cmidrule(lr){2-10}
 & \textcolor{TrustOrange}{\textbf{DA-DMD}} & \makecell{61.9\\[-2pt]\dmdgain{-0.6}} & \makecell{\textbf{59.7}\\[-2pt]\dmdgain{+0.8}} & \makecell{79.9\\[-2pt]\dmdgain{-1.0}} & \makecell{\textbf{78.8}\\[-2pt]\dmdgain{+0.1}} & \makecell{67.0\\[-2pt]\dmdgain{-1.2}} & \makecell{67.1\\[-2pt]\dmdgain{-0.3}} & \makecell{\textbf{66.8}\\[-2pt]\dmdgain{+0.4}} & \makecell{\textbf{68.8}\\[-2pt]\dmdgain{+1.2}} \\
\midrule
\multirow{2}{*}{SD3.5-L} & DMD2 & 64.0 & 57.0 & 80.1 & 75.6 & 65.1 & 64.4 & 63.5 & 62.8 \\
\cmidrule(lr){2-10}
 & \textcolor{TrustOrange}{\textbf{DA-DMD}} & \makecell{\textbf{66.4}\\[-2pt]\dmdgain{+2.4}} & \makecell{\textbf{60.3}\\[-2pt]\dmdgain{+3.3}} & \makecell{\textbf{80.8}\\[-2pt]\dmdgain{+0.7}} & \makecell{\textbf{78.5}\\[-2pt]\dmdgain{+2.9}} & \makecell{\textbf{68.0}\\[-2pt]\dmdgain{+2.9}} & \makecell{\textbf{65.8}\\[-2pt]\dmdgain{+1.4}} & \makecell{\textbf{66.6}\\[-2pt]\dmdgain{+3.1}} & \makecell{\textbf{65.4}\\[-2pt]\dmdgain{+2.6}} \\
\bottomrule
\end{tabular}
}%
\end{table*}

\begin{figure*}[t]
\begin{center}
\includegraphics[width=1.\linewidth]{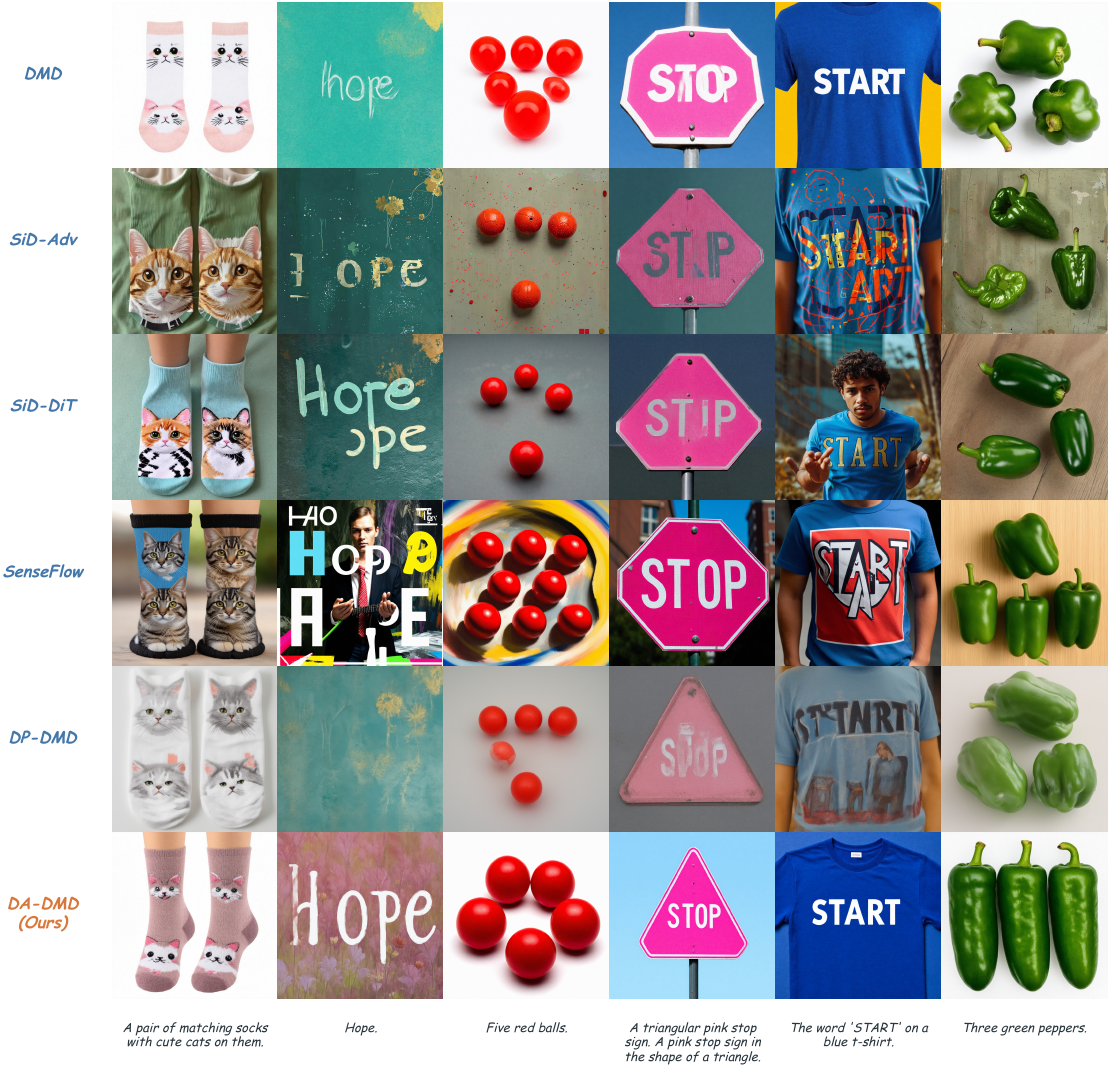}
\caption{
\textbf{Qualitative comparison with representative distillation baselines on SD3.5-M.}
DA-DMD demonstrates stronger prompt adherence and visual consistency across challenging cases involving counting, text rendering, fine-grained attributes, and object identity, compared with DMD, SiD-Adv, SiD-DiT, SenseFlow, and DP-DMD.
}
\label{fig:sd35m_baseline}
\end{center}
\end{figure*}

\end{document}